\documentclass[journal,trans]{IEEEtran}

\ifCLASSINFOpdf
\else
\fi

\usepackage{epsfig} 
\usepackage{cite} 
\usepackage{algorithm,algpseudocode,algorithmicx}
\usepackage[mathscr]{eucal}
\usepackage{amsmath,amsfonts,amssymb}
\usepackage{graphicx,subfigure,cases,multirow,graphics}
\usepackage[colorlinks,citecolor=blue,urlcolor=black,linkcolor=red]{hyperref}
\usepackage[displaymath,mathlines,switch]{lineno}
\usepackage{booktabs}
\usepackage{threeparttable}

\newlength\savewidth

\newlength\savedwidth

\newcommand{\bS}{\mathbb S}
\newcommand{\bR}{\mathbb R}
\newcommand{\bZ}{\mathbb Z}
\newcommand{\bM}{\mathbb M}
\newcommand{\bD}{\mathbb D}

\newcommand{\rC}{\mathscr C}

\newcommand{\cC}{\mathcal C}

\newcommand{\cL}{\mathcal L}
\newcommand{\cF}{\mathcal F}

\newcommand{\cD}{\mathcal D}
\newcommand{\cU}{\mathcal U}
\newcommand{\cG}{\mathcal G}
\newcommand{\cV}{\mathcal V}
\newcommand{\cE}{\mathcal E}
\newcommand{\cS}{\mathcal S}
\newcommand{\cP}{\mathcal P}

\newcommand{\kC}{\mathfrak C}

\newcommand{\kL}{\mathfrak L}

\newcommand{\fb}{\mathbf b}
\newcommand{\fp}{\mathbf p}
\newcommand{\fs}{\mathbf s}
\newcommand{\fe}{\mathbf e}
\newcommand{\fu}{\mathbf u}
\newcommand{\fv}{\mathbf v}
\newcommand{\fn}{\mathbf n}
\newcommand{\fx}{\mathbf x}
\newcommand{\fy}{\mathbf y}

\newcommand{\fI}{\mathbf I}

\DeclareMathOperator\Dist{Dist}
\DeclareMathOperator\Lip{Lip}
\DeclareMathOperator\Length{Length}
\begin{document}
%
\title{Grouping Boundary Proposals for Fast Interactive Image Segmentation}


\author{\IEEEauthorblockN{Li Liu, Da Chen, Minglei Shu and Laurent D. Cohen,~\IEEEmembership{Fellow,~IEEE}}
\thanks{Li~Liu, Minglei Shu and Da~Chen are with Shandong Artificial Intelligence Institute, Qilu University of Technology (Shandong Academy of Sciences),  250014 Jinan, China.}  
\thanks{Laurent~D.~Cohen is with University Paris Dauphine, PSL Research University, CNRS, UMR 7534, CEREMADE, 75016 Paris, France.}
}

\maketitle

\markboth{Journal of \LaTeX\ Class Files,~Vol.~, No.~, JUNE~2022}%
{Shell \MakeLowercase{\textit{et al.}}: Bare Demo of IEEEtran.cls for Journals}
%



\IEEEtitleabstractindextext{%
\begin{abstract}
Geodesic models are known as an efficient tool for solving various image segmentation problems. Most of existing approaches only exploit local pointwise image features to track geodesic paths for delineating the objective boundaries. However, such a segmentation strategy cannot take into account the connectivity of the image edge features, increasing the risk of shortcut problem, especially in the case of complicated scenario. In this work, we introduce a new image segmentation model based on the minimal geodesic framework in conjunction with an adaptive cut-based circular optimal path computation scheme and a graph-based boundary proposals grouping scheme. Specifically, the adaptive cut can disconnect the image domain such that the target contours are imposed to pass through this cut only once. The boundary proposals are comprised of precomputed image edge segments, providing the connectivity information for our segmentation model. These boundary proposals are then incorporated into the proposed image segmentation model, such that the target segmentation contours are made up of a set of selected boundary proposals and the corresponding geodesic paths linking them. 
Experimental results show that the proposed model indeed outperforms state-of-the-art minimal paths-based image segmentation approaches.   
\end{abstract}

}

\IEEEdisplaynontitleabstractindextext

%
\IEEEpeerreviewmaketitle

\section{Introduction}
Image segmentation is a fundamental and challenging issue in image analysis, whose main goal is to extract the boundary of target regions. Energy minimization models have been widely applied to a broad variety of segmentation problems. Among them,  the interactive segmentation strategy allows to incorporate user interaction into the optimization process, yielding promising results. The interactions are applied either to provide necessary initialization for the segmentation models or to generate efficient constraints in the course of the image segmentation procedures.

Scribbles are very often considered as manual interaction in image segmentation models.  A common way is to require the user to tag some pixels respectively as either foreground or background labels. Many graph-based models use such type of input, such as the graph cut-based models~\cite{boykov2006graph,vicente2008graph,couprie2010power}, the random walk-based models~\cite{grady2006random,yang2010user,zhang2010diffusion}, lazy snapping algorithm~\cite{li2004lazy} and the distance-based models~\cite{arbelaez2004energy,bai2009geodesic,price2010geodesic,chen2018fast}. The regions of interest are extracted by minimizing the energy functional defined on the weighted graph. In addition to providing initial location of target regions, the scribbles associated to different image regions are also applied to extract necessary features, which are regarded as statistical priors to define a data-driven  term~\cite{nguyen2012robust,spencer2018parameter,gao20123d}. Besides, the scribbles can also be exploited to refine the segmentation results. An interesting example is the deep learning-based interactive segmentation method~\cite{wang2019deep}. The user interactions indicate the mis-segmentation regions, which are regarded as hard constraints to correct the segmentation. Another simple interaction paradigm is constructed by dragging a rectangle box around the desirable object~\cite{rother2004grabcut,lempitsky2010image} aiming to indicate the background region. Final results detected by these models also depend on the discrete graph-based optimization scheme. However, scribble-based corrections are usually needed to refine the segmentation in many cases, where the interaction dose not provide sufficient prior information. 

In contrast to using scribbles which carry regional information to create initialization or constraints, boundary-based interactions are also taken into account in many segmentation algorithms.  Placing an initial contour close to the target boundary is a  widely considered way for initializing active contours, especially for  models implemented with a parametric curve evolution scheme or a level set scheme~\cite{kass1988snakes,xu1998snakes,li2008minimization,brox2009local,cai2021avlsm}. The segmentation can be achieved by evolving initial contours through the forces derived from the image data and adequate geometric properties. An alternative way is to take a set of landmark points as boundary-based interactions in many segmentation models. Specifically, for the active contour approaches such as~\cite{badshah2010image,mabood2016active}, the landmark points are supposed to be exactly placed at the target boundary, and serve as attractors which encourage the evolutional contours to move towards these points. Windheuser~\emph{et al.}~\cite{windheuser2009beyond} proposed a graph-based interactive segmentation model, for which the initialization is a set of ordered imprecise landmark points. In this model, all the landmark points allow to distribute along a shifted contour to the target boundary. 

Shortest paths-based segmentation models exploit a series of optimal curves associated to adequate energy functionals, tracked in either a discrete domain~\cite{miranda2012riverbed,mortensen1998interactive,dijkstra1959note} or a continuous one~\cite{cohen1997global}, to delineate the target boundaries. The landmark points define the source point as well as the end point for each optimal curve. Among these models, the minimal path model based on the Eikonal partial differential equation (PDE) has shown their ability in extracting  structures of interest in different complex scenarios, benefiting from the fast numerical solution schemes and the global optimal characteristic~\cite{peyre2010geodesic}. In its basic formulation, the geodesic path is an open curve that globally minimizing the weighted length function between two user-provided endpoints. In practical application, the original framework needs to be modified to extract closed curves composing the boundary of the target region. Different variants of minimal path algorithms have been devised to extract target regions with a priori knowledge.  In the remaining of this section, we briefly introduce the existing minimal paths-based segmentation models and give the motivation of our work.

\subsection{Segmentation Models based on Geodesic Paths}
\label{subsec_ReviewMPs}

In image analysis, it is a fundamental task to detect closed contours to delineate the target boundaries. Many segmentation approaches have been devoted to address the problem of building simple closed contours using a family of piecewise geodesic paths~\cite{peyre2010geodesic}.  Models relying on saddle points~\cite{cohen1997global} or  keypoints~\cite{benmansour2009fast} can detect a simple closed contour starting from a single source point.  In~\cite{mille2015combination,chen2017global,cohen2001multiple,chen2019region}, a set of prescribed points exactly placed at the target boundary are leveraged for model initialization, allowing the user to incorporate more flexible intervention into the segmentation processing. 
The classical circular geodesic method~\cite{appleton2005globally} is an alternative way for closed contour detection, which treats each contour as a simple closed minimal path. In its basic setting, only a single point inside the target region is required to initialize the algorithm. However, the classical circular geodesic model applied an axis cut such that a particular  domain is taken as the search space to handle shapes which violate a cut-convexity assumption~\cite{appleton2005globally}, increasing the computational time. Chen~\emph{et al}.~\cite{chen2021geodesic} proposed to combine two geodesic paths as a closed contour, in which the region-based homogeneity features are also taken into account for image segmentation. 

\begin{figure}[t]
\centering
\includegraphics[width=0.95\linewidth]{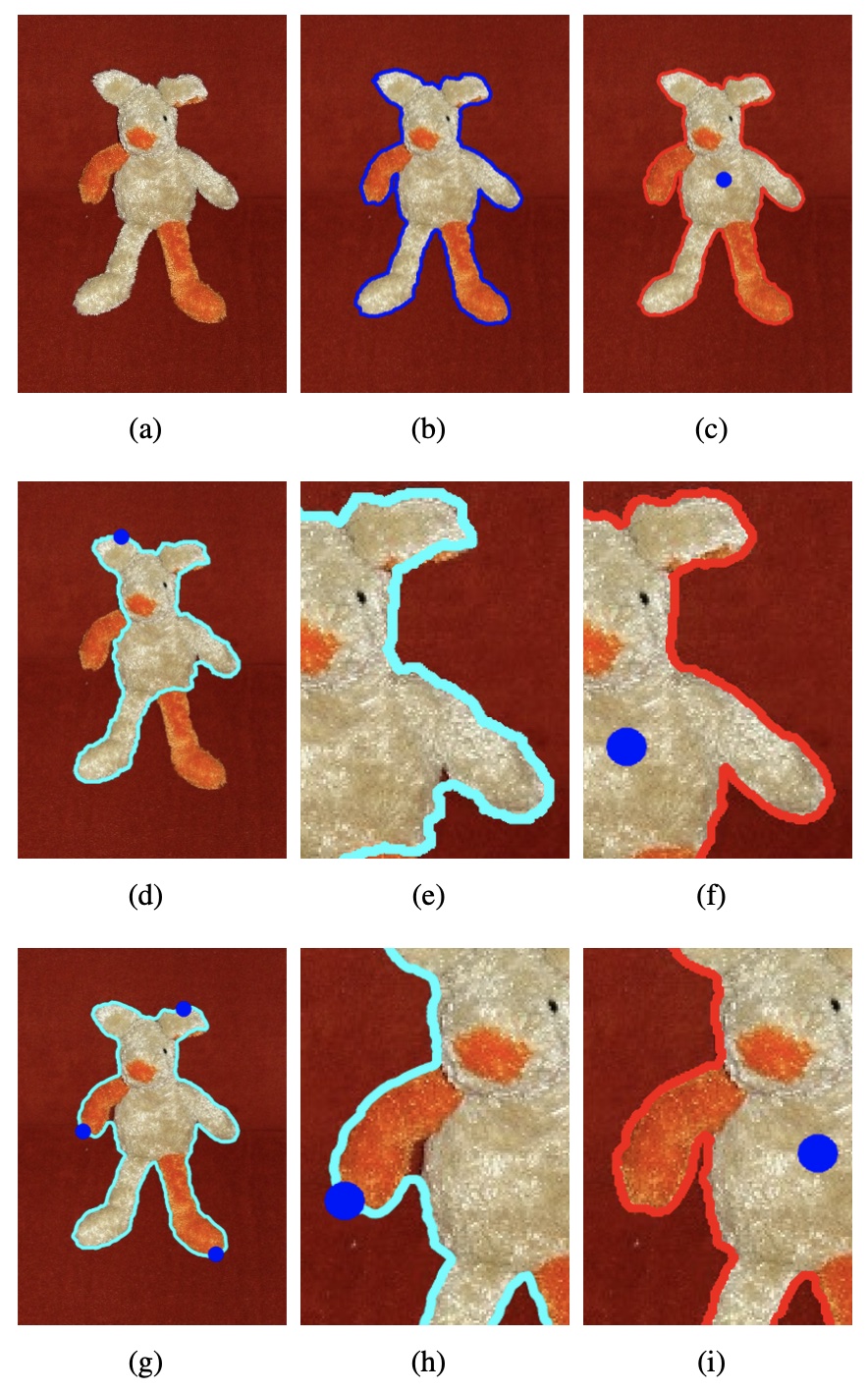}
\caption{Examples for illustrating the advantages of the proposed image segmentation model. (\textbf{a}) An original image sampled from the Grabcut dataset. (\textbf{b}) The ground truth denoted by a blue line. (\textbf{c}), (\textbf{d}) and (\textbf{g}) Segmentation results obtained via the proposed model, the growing minimal path  model~\cite{benmansour2009fast} and the combination of piecewise-geodesic paths model~\cite{mille2015combination}, respectively. (\textbf{e}) and (\textbf{h}) Close-up view of the segmentation contours of (\textbf{d}) and (\textbf{g}), respectively. (\textbf{f}) and (\textbf{i}) Close-up view of the segmentation contour of (\textbf{c}). }
\label{fig:examples}	
\end{figure}

Unfortunately, existing segmentation models based on geodesic paths do not take the advantages of edge connectivity enhancement. In order to solve this issue, we propose a new geodesic segmentation model relying on an adaptive cut, which feature edge connectivity derived from precomputed boundary proposals. It is designed for fast image segmentation by providing an arbitrary landmark point within the target object region. The entire segmentation model is made up of two steps: (i) discrete graph construction and (ii) Interactive segmentation, as summarized in Algorithm~\ref{algo_Overview}. Furthermore, the step of discrete graph construction does not require user interaction, thus can be categorized as an offline step. This step prepares the necessary data which will be submitted to next step for addressing the interactive segmentation task.
Examples for illustrating the advantages of the proposed  model can be seen in Fig.~\ref{fig:examples}. The shortcuts problems occur in the minimal path growing model~\cite{benmansour2009fast} and the combination of piecewise-geodesic paths model~\cite{mille2015combination} as indicated in Figs.~\ref{fig:examples}d and~\ref{fig:examples}g. Our model enables to blend the benefits of geodesic paths and the precomputed  boundary proposals. In our model a simple closed contour is constructed via a perceptual grouping scheme, implemented through the Dijkstra's algorithm~\cite{dijkstra1959note} over a graph. In the existing perceptual grouping method~\cite{wang2013interactive}, the prescribed trajectories representing tubular centerlines are introduced to use for tubular structure tracking, by grouping the trajectories those belonging to the same structure as the shortest path. They build the graph by exploiting a straight segment to connect two neighbouring trajectories and utilizing its Euclidean length to measure the connection cost, which may lead to approximation errors. In our previous work~\cite{liu2021new,liuTractory2022}, we propose a more reasonable way to link neighbouring trajectories by curvature-penalized geodesic paths, and the weight for the edge is established based on geodesic curvature instead of image features. However, those grouping works are optimized for detecting open curve between two endpoints. 
In this work, our grouping method aims at closed contour extraction with one user-provided point. The contribution of the paper is twofold.
\begin{itemize} 
\item Firstly, the connectivity enhancement is incorporated into the minimal paths-based segmentation framework. We establish a new minimal paths-based image segmentation method in conjunction with prescribed boundary proposals. With the contour closure assumption, the final segmentation is achieved by selecting an ordered sequence of boundary proposals connected by minimal paths, yielding a simple closed contour.
\item Secondly, the axis cut method used in the classical circular geodesic model~\cite{appleton2005globally} is extended to the case of adaptive cut regarded as a geodesic path linking a given landmark point and the image domain boundary. Furthermore, we introduce a reasonable computation method encouraging the adaptive cut to intersect with the target boundary only \emph{once}, leading to the possibility of finding circular paths from a source point located at this adaptive cut. 
\end{itemize}
The remaining of this manuscript is organized as follows. In Section~\ref{sec:MPs}, we briefly introduce the background on the classical geodesic path model and on the computation of image edge appearance features. Sections~\ref{sec_MainModel} and~\ref{sec:BG} formulate the main contributions in this work. Specifically, the construction of the adaptive cut is presented in Section~\ref{sec_MainModel}. Section~\ref{sec:BG} present the details on the construction of the discrete graph and the interactive segmentation procedure.  The experimental results and the conclusion are presented in Sections~\ref{sec:Exp} and \ref{sec:conclusion}, respectively.  

\section{Background}
\label{sec:MPs}
\subsection{Cohen-Kimmel Minimal Path Model}
Let $\Omega\subset\bR^n~(n=2,3)$ be an open and bounded image domain with dimension $n$. To detect the target structure in the given image, Cohen and Kimmel~\cite{cohen1997global} introduced a minimal path model which delineates target boundaries via minimizing curves of an energy. This original  model applies a  scalar-valued function $\psi:\Omega\to\bR^+$ to define the energy of a Lipschitz continuous curve $\gamma:[0,1]\to\Omega$, which reads as
\begin{equation}
\label{eq:IsoEnergy}
\cL(\gamma):=\int_0^1 \psi(\gamma(u))\|\gamma^\prime(u)\|du,
\end{equation}
where $\gamma^\prime=d\gamma/du$ is the first-order derivative of $\gamma$, and $\psi:\Omega\rightarrow\bR^+$ is a  function featuring lower values near desired features, which is also referred to as \emph{potential}.
In the basic formulation of the Cohen-Kimmel model~\cite{cohen1997global}, a boundary proposal between two points $\fs$ and $\fx$ is modeled as a geodesic path $\cG_{\fs,\fe}$ that has minimal weighted length~\eqref{eq:IsoEnergy} among all Lipschitz paths linking from the source point $\fs$ and end point $\fx$, i.e.
\begin{equation} 
\label{eq:GeodesicPath}
\cG_{\fs,\fx}=\underset{\gamma\in \text{Lip}([0,1],\Omega)}{\arg\min}\lbrace\cL(\gamma)\mid\gamma(0)=\fs,\gamma(1)=\fx\rbrace,
\end{equation}
where $\text{Lip}([0,1],\Omega)$ represents the set of all Lipschitz continuous curves $\gamma:[0,1]\to\Omega$. A crucial ingredient for tracking a geodesic path emanating from $\fs$ lies at the definition of a geodesic distance map, or a minimal action map, $\cU_\fs:\Omega\to\bR^+_0$ such that
\begin{equation} 
\label{eq:GeodesicDistance}
\cU_{\fs}(\fx)=\underset{\gamma\in \text{Lip}([0,1],\Omega)}{\inf}\{\cL(\gamma)\mid\gamma(0)=\fs,\,\gamma(1)=\fx\}.
\end{equation}
As discussed in~\cite{cohen1997global}, the geodesic distance map $\cU_\fs$ admits the following Eikonal equation 
\begin{equation} 
\label{eq:EikonalPDE}
\|\nabla\cU_\fs(\fx)\|=\psi(\fx),~~\forall \fx\in\Omega\backslash\{\fs\},
\end{equation}
with boundary condition $\cU_\fs(\fs)=0$, where $\nabla\cU_\fs$ denotes the standard  Euclidean gradient of $\cU_\fs$. 

The isotropic Eikonal equation can be solved effectively by the classical Fast Marching  method~\cite{sethian1999fast}. 
Once the geodesic distance map is estimated, a parameterized curve $\cG_{\fx,\fs}$ from a target point $\fx\in\Omega$ to the fixed source point $\fs$ can be tracked through the solution to the following gradient descent ODE on the distance map, i.e. 
\begin{equation}
\label{eq:ODE}
\cG_{\fx,\fs}^\prime(u)=-\frac{\nabla\cU_\fs(\cG_{\fx,\fs}(u))}{\|\nabla\cU_\fs(\cG_{\fx,\fs}(u))\|},
\end{equation}
with $\cG_{\fx,\fs}(0)=\fs$. The numerical scheme for~\eqref{eq:ODE} is terminated till the source point $\fs$ is reached.  Then the geodesic path $\cG_{\fs,\fx}$ parameterized by its arc-length can be retrieved by reversing the path $\cG_{\fx,\fs}$ with $\cG_{\fs,\fx}(0)=\fs$ and $\cG_{\fs,\fx}(1)=\fx$.

\subsection{Edge Appearance Features}
The edge appearance features  of a color image $\fI=(I_1,I_2,I_3):\Omega\to\bR^3$ characterize the likelihood of a point belonging to an image edge. Let $J:\Omega\to\bR^+_0$ be a map that defines the edge appearance features. In general, $J$ can be constructed using the image gradients as follows~\cite{sochen1998general,chen2021geodesic}:
\begin{equation} 
\label{eq:EdgeAppearanceFeature}
J(\fx)=\sqrt{\sum_{k=1}^n\Big(\Arrowvert(\partial_xG_\sigma\ast I_k)(\fx)\Arrowvert^2+\Arrowvert(\partial_yG_\sigma\ast I_k)(\fx)\Arrowvert^2\Big)},
\end{equation}
where $\ast$ is the convolution operator, $G_\sigma$ represents the Gaussian kernel with standard derivation $\sigma$, $\partial_x$ and $\partial_y$ denote the partial derivative along x- and y-axis directions, respectively. 
In practice, we normalize the magnitudes $J$ to the range $[0,1]$ to generate a new appearance feature function $g$ by 
\begin{equation} 
\label{eq:EdgeAppearanceFeatureNormalized}
g(\fx)=\frac{J(\fx)}{\|J\|_\infty},\quad\forall\fx\in\Omega.
\end{equation}
A strong value of $g(x)$ indicates that a high possibility that $x$ is an edge point. In the context of image segmentation, the construction for the potential $\psi$, see~Eq.\eqref{eq:IsoEnergy} can be naturally built by the function $g$.

\section{Adaptive Cut for Circular Optimal Paths}
\label{sec_MainModel}
\subsection{Overview}
The circular geodesic model~\cite{appleton2005globally} provides an avenue for computing closed geodesic paths using a single point inside the target contour. 
In its basic setting, the axis cut emanating from the landmark point $\fp$ to $\partial\Omega$, is imposed to align the direction of  an axis. This model is suitable for the case that the target regions have a cut-convexity shape, limiting its potential applications in practice. In order to solve this problem, we introduce a more flexible strategy for finding circular optimal paths, relying on an adaptive cut which intersects the target boundary exactly once. The construction of the adaptive cut lies at the first stage of the \emph{interactive segmentation step} in the propose model, as presented in Line~\ref{algLine_AdaptiveCut} of Algorithm~\ref{algo_Overview}.

\begin{algorithm}[!t]
\caption{Summary Algorithm of Our Model}	
\label{algo_Overview}
\begin{algorithmic}
\renewcommand{\algorithmicrequire}{\textbf{Input:}}
\renewcommand{\algorithmicensure}{\textbf{Output:}}
\Require A landmark point $\fp\in\Omega$ inside the target region.
\Ensure  Segmentation contour.
\renewcommand{\algorithmicrequire}{\textbf{Discrete Graph Construction Step:}}
\Require  
\end{algorithmic}
\begin{algorithmic}[1]
\State Detect a set of boundary proposals $\cS_i$ for $i=1,2,\cdots$.
\State Construct a discrete graph $G$ by computing its edge set and edge weights.
\renewcommand{\algorithmicrequire}{\textbf{Interactive Segmentation Step:}}
\Require 
\State Compute a structure-aware adaptive cut $\rC_{\fp}$ linking from the landmark $\fp$ to a point at the domain boundary.
\label{algLine_AdaptiveCut}
\State Construct a set $\chi_\fp$ involving all admissible circular paths.
\State Find a circular path $\cC^*$ of minimum energy from the set $\chi_\fp$, which is taken as the final segmentation contour.
\end{algorithmic}
\end{algorithm}

\subsection{Construction of A Set of Boundary Proposals}
In our model, a boundary proposal is regarded as a continuous curve $\cS_i\subset\Omega$, each of which characterizes a piece of an image edge.  As an example, we visualize in Fig~\ref{fig:cutExamples}b  the boundary proposals via with colored lines, computed from a synthetic as depicted in Fig.~\ref{fig:cutExamples}a. 

In numerical implementations, the construction of the boundary proposals $\cS_i$ can be carried out through many existing edge detectors~\cite{canny1986computational,arbelaezContour2011,jacob2004design}. The detection of edge structures is implemented in a discretization grid $\bM_h=h\bZ^2\cap\Omega$ where $h$ is the grid scale. These edge structures of one grid point width might be non-simple with multiple branches connecting to junction points, where each junction point is defined as an edge point which has more than two neighbour points. Then we remove all the junction points from the detected edge structures to generate a series of disjoint edge segments, i.e. the boundary proposals. We also remove small fragments\footnote{In other words, the  boundary proposals whose grid points are less than a given threshold value are removed.} to reduce the computation complexity of the model. 

\subsection{Construction of A Structure-aware Adaptive Cut}
\label{sub_GeodesicCut}
In our model, an adaptive cut is regarded as a geodesic path connecting a \emph{landmark} point $\fp$ inside the target region and a detected point lying at the domain boundary $\partial\Omega$. It is built partially relying on the boundary proposals $\cS_i$. As in Section~\ref{sec:MPs}, a key step for computing an appropriate adaptive cut is the construction of the potential $\psi$, see Eq.~\eqref{eq:IsoEnergy}. 

The boundary proposals provide the image edge information. Thus we utilize these boundary proposals to compute a structure-aware potential $\psi:=\psi_{\rm str}$
\begin{equation} 
\label{eq:PotentialFunction}
\psi_{\rm{str}}(\fx):=\phi(\fx)\delta(\fx),
\end{equation}
where  $\phi:\Omega\to\bR_0^+$ is an edge indicator and where   $\delta:\Omega\to\{1,\infty\}$ is a boundary proposal indicator.
Specifically, the edge indicator $\phi$ is defined as
\begin{equation} 
\label{eq:EdgeDetector}
\phi(\fx)=\exp\big(\tau g(\fx)\big)+\tilde{w},
\end{equation}
where $\tau$ and $\tilde{w}$ are two positive constants, and $g$ is the magnitude  of the image gradient, see Eq.~\eqref{eq:EdgeAppearanceFeatureNormalized}.
The function $\delta$ associated to the boundary proposals $\cS_i$ is expressed as
 \begin{equation}
\label{eq:TopologyIndicator}
\delta(\fx) =  
\begin{cases}
+\infty ,&\text{if~}\fx\in\cup\cS_i, \\
1,& \text{otherwise}.
\end{cases}
\end{equation}

Then one can estimate a geodesic distance map $\cU_\fp:\Omega\to\bR^+_0$ by solving the classical Eikonal PDE~\eqref{eq:EikonalPDE} with boundary condition $\cU_\fp(\fp)=0$, by setting $\psi(\fx)=\psi_{\rm{str}}(\fx)$. The endpoint  $\fb\in\partial\Omega$ of the target adaptive cut $\rC_\fp$  can be detected by finding a point of minimum distance value, i.e.
\begin{equation}
\fb:=	\underset{\fx\in\partial\Omega}{\arg\min}~\cU_{\fp}(\fx).
\end{equation}
Finally, the adaptive cut $\rC_{\fp}$ is then can be generated using the solution to the gradient descent ODE~\eqref{eq:ODE} on $\cU_\fp$ such that $\rC_{\fp}(0)=\fp$ and $\rC_{\fp}(1)=\fb$.

\begin{figure*}[t]
\centering
\includegraphics[width=0.98\linewidth]{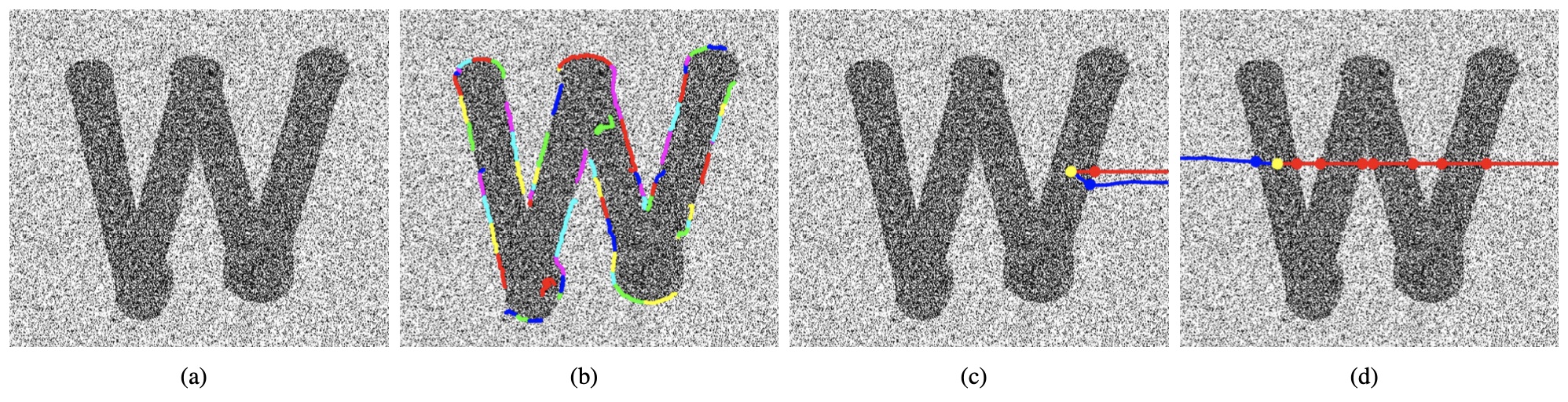}
\caption{An example for visualizing the difference between the axis cut and the proposed adaptive cut. (\textbf{a}) A synthetic image. (\textbf{b}) Boundary proposals superimposed on the original synthetic image. (\textbf{c})  (\textbf{d}) The axis cut~\cite{appleton2005globally} and adaptive cut are represented by the red and blue lines.  The yellow points are the landmark  points inside the target region. The red (resp. blue) dots are the interaction points between the target contour and the axis cuts (resp. adaptive cuts).}
\label{fig:cutExamples}	
\end{figure*}

\begin{figure*}[t]
\centering
\includegraphics[width=0.98\linewidth]{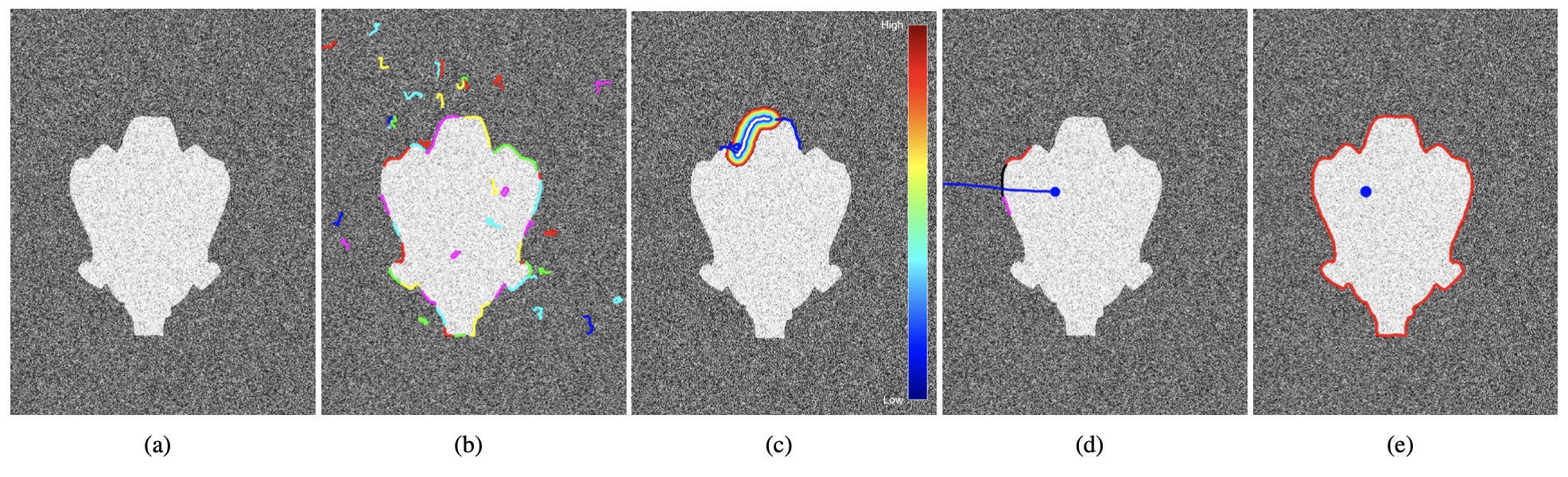}
\caption{Overview of the introduced boundary proposal grouping method. (\textbf{a}) A synthetic image. (\textbf{b}) The boundary proposals superimposed on the synthetic image. (\textbf{c}) The construction for the neighbourhood $\Psi_i$ of the boundary proposal $\cS_i$  indicated by a white line. The neighbourhood  $\Psi_i$ is done by thresholding the geodesic distances superimposed on the synthetic image. Blue and red lines represent the neighboring boundary proposals of the chosen one. Note that in this figure only distance values for the points which satisfy that $\cU_i(\fx)\leq\eta$ are visualized.  (\textbf{d}) The extraction of the discrete shortest path comprised of sequential boundary proposals. The blue dash line is the adaptive cut. The boundary proposals indicated by red and magenta crepresent the nodes $\vartheta_i$ and $\vartheta_j$ whose connecting geodesic path (black solid line) intersects with the adaptive cut. (\textbf{e}) The red line is the final contour.}
\label{fig:procedure}	
\end{figure*}

One can point out that the computed adaptive cut $\rC_{\fp}$ will not pass through the boundary proposals due to the use of the indicator $\delta$. Moreover, the potential $\psi_{\rm str}$ has high values around the image edges, and low values otherwise. Such a setting encourages the adaptive cut $\rC_{\fp}$ to pass through the target boundary just once. In Figs.~\ref{fig:cutExamples}c and~\ref{fig:cutExamples}d, two examples are exploited to exhibit the difference between the axis cuts~\cite{appleton2005globally} (red lines) and the proposed adaptive cuts (blue lines). One can see that the axis cut in Fig.~\ref{fig:cutExamples}c intersects with the target boundary only once, which is also the case for the proposed adaptive cut. In Fig.~\ref{fig:cutExamples}d, the axis cut has multiple intersection points with the target boundary. In this case, one needs to perform a complicated scheme to track a circular geodesic path in junction with such an axis cut. In contrast, the proposed adaptive cut has only one intersection point even through the target regions have very complicated shapes. 

Given a source point $\fs$ located at the adaptive cut,  one can track a circular optimal path passing through the source point $\fs$, implemented using a similar scheme to the classical circular geodesic model~\cite{appleton2005globally} in the case that the axis cut intersects with the target boundary just once, demonstrating the advantages of using the adaptive cut.
However, such a circular optimal path tracking scheme in conjunction  with an adaptive cut cannot take into account the precomputed boundary proposals  carrying very useful information (e.g. edge connectivity enhancement) for segmentation. In order to avoid this issue, we propose a new method for extracting circular optimal paths, where the procedure is enhanced by the precomputed boundary proposals.  

\section{Grouping Boundary Proposals for Fast Image Segmentation}
\label{sec:BG}
In this section, we present a fast and reliable method for extracting circular optimal paths for interactive segmentation based on an adaptive cut computation scheme and a perceptual grouping scheme using boundary proposals. The proposed model relies on the computation of discrete optimal paths from a graph, depending on the geodesic distances derived from adequately constructed geodesic metrics. 

The proposed model searches for closed contours as  circular optimal paths from a graph $G=(\cV,\cE)$, where $\cV$ is a set of $N$ nodes, and  $\cE$ is a set of edges connecting two adjacent nodes in $\cV$. As considered in~\cite{wang2013interactive,liu2021new}, a node $\vartheta_i\in\cV$ of the graph represents a  boundary proposal $\cS_i$ in the image domain. Two nodes $\vartheta_i$ and $\vartheta_j$ with $i\neq j$ are said to be adjacent if  the  corresponding boundary proposals $\cS_i$ and $\cS_j$ are neighbours. We denote by $e_{i,j}\in\cE$ the edge that links the node $\vartheta_i$ to an adjacent one $\vartheta_j$. Also, a scalar-valued weight $w_{i,j}>0$ is  assigned to each edge $e_{i,j}$, which can be estimated using geodesic models, see Section~\ref{subsec_EdgeWeight}. In the remaining of this section, we present main components of the proposed model, involving the construction of the edge set $\cE$ and the computation for each edge weight, introduced in Sections~\ref{sub_CutSegment} and~\ref{subsec_EdgeWeight}, respectively.
In Fig.~\ref{fig:procedure}, we illustrate the pipeline  of the proposed boundary proposal grouping model, where Fig.~\ref{fig:procedure}a shows a synthetic image and where Fig.~\ref{fig:procedure}b visualizes the boundary proposals using different colors. In the remaining of this section, we introduce the details for each step.
 
\subsection{Construction of the Edge Set for the Discrete Graph}
\label{sub_CutSegment}
In the proposed model, the discrete graph encoding image edge features is generated using the boundary proposals $\cS_i$ for $i=1,2,\cdots$. The basic idea is to take each boundary proposal $\cS_i$ as a node $\vartheta_i$ of the graph $G$.  Along this idea, we introduce a method to identify all the adjacent boundary proposals for each fixed boundary proposal $\cS_i$, providing that an appropriate adaptive cut $\rC_{\fp}$ is given. Note that the construction of the graph does not rely on the user interaction, as presented in Algorithm~\ref{algo_Overview}.
\subsubsection{Finding Adjacent Boundary Proposals}
For a fixed boundary proposal $\cS_i$, the detection of its adjacent boundary proposals $\cS_j$ relies on the construction of the tubular neighbourhood $\Psi_i\subset\Omega$ that surrounds $\cS_i$. A widely considered idea for computing $\Psi_i$ is to leverage the Euclidean distance map associated to $\cS_i$, as in~\cite{wang2013interactive,liu2021new,chen2019region}. However, the Euclidean distances-based neighbourhood may involve some boundary proposals that not belong to the target boundary, thus usually identifying unexpected adjacent boundary proposals. In order to handle this issue, we  take into account the image data for the construction of the tubular neighbourhood. In other words, we propose to adopt a geodesic distance map $\cU_i:\Omega\to\bR^+_0$  (s.t. $\cU_i(\fx)=0$, $\forall \fx\in\cS_i$) for $\Psi_i$ with respect to an adequately designed geodesic metric. The computation of $\cU_i$ is presented in next section.
The neighbourhood $\Psi_i$ of the boundary proposal $\cS_i$ is determined by thresholding the distance map  $\cU_i$ with a value $\zeta\in\bR^+$ as
\begin{equation}
\label{eq:Neighbourhood}
\Psi_i:= \{\fx\in\Omega\mid\cU_i(\fx)\leq \zeta\}. 
\end{equation}
A boundary proposal $\cS_j$ for any $j\neq i$ is said to be \emph{adjacent} to $\cS_i$  if $\cS_j\cap\Psi_i\neq \emptyset$ is satisfied. 

\subsubsection{Estimating the Geodesic Distance Map $\cU_i$}
In our model, the geodesic distance maps $\cU_i$ for constructing tubular neighbourhoods of the boundary proposals $\cS_i$ is computed using the classical  isotropic Riemannian metrics, involving a potential defined by
\begin{equation}
\label{eq_SegPotential}
\psi_{\rm{Seg}}(\fx)=\frac{1}{\phi(\fx)+\varepsilon}
\end{equation}
where  $\varepsilon\in\bR^+$ is  a constant. As in the classical Cohen-Kimmel model, each geodesic distance map  $\cU_i$ admits the following isotropic Eikonal PDE
\begin{equation} 
\label{eq:newEikonalPDE}
\begin{cases}
\|\nabla\cU_i(\fx)\|=\psi_{\rm{Seg}}(\fx),&\forall \fx\in\Omega\backslash\cS_i\\
\cU_i(\fx)=0,&\forall \fx\in\cS_i.
\end{cases}
\end{equation}
The computation of geodesic distances $\cU_i$ w.r.t. $\cS_i$ are implemented using the classical fast marching method~\cite{sethian1999fast}. The computation procedure is terminated immediately  once a point $\fx\in\Omega$ such that $\cU_i(\fx)\geq\eta$ has been reached by the geodesic distance front. We illustrate an example for such an estimated geodesic distance map $\cU_i(\fx)$ in Fig~\ref{fig:procedure}c. The values of $\cU_i$ increase slowly along the regions featuring  high values of $\phi$ which usually indicate image edges, and increase fast along low values of $\phi$ corresponding to flat regions. This means that the tubular neighbourhood shapes are partially determined by the image edge features.

\subsection{Computation of Connection Paths and Edge Weights}
\label{subsec_EdgeWeight}
The graph to construct is denoted by $G=(\cV,\cE)$ where $\cV$ and $\cE$ are the sets of nodes and edges, respectively. In this section, we build the edge weights for each graph edge  $w_{i,j}$ assigning to each edge $e_{i,j}\in\cE$. In our work, we estimate the weight $w_{i,j}$ as the geodesic distance between the boundary proposals $\cS_i$ and $\cS_j$. For this purpose, we briefly introduce a general metric $\cF:\bM\times\bR^n$ ($n=2,3$) to measure the weighted length of a smooth curve, where $\bM\subset\bR^n$ is an open bounded and connected domain. At each point $\fx\in\bM$, the metric $\cF(\fx,\fu)$ is defined as an \emph{asymmetric} norm of the vector $\fu$ over the tangent space $\bR^n$. In our model, we consider the classical isotropic metric~\cite{cohen1997global} and the curvature-penalized metrics~\cite{chen2017global,duits2018optimal,mirebeau2018fast} as instances of $\cF$, see Appendix~\ref{subsec_Convexity} for the detailed construction of these metrics. 

\subsubsection{Connection Path between Two Adjacent Boundary Proposals}
By the metric $\cF$, the weighted curve length of a curve $\gamma\in\Lip([0,1],\bM)$ can be formulated as
\begin{equation}
\label{eq_FinslerGeoEnergy}
\cL_\cF(\gamma)=\int_0^1\cF(\gamma(u),\gamma^\prime(u))du.
\end{equation}
The goal is to compute an optimal path $\cG_{i,j}$ which connects the boundary proposal $\cS_i$ to its adjacent one $\cS_j$. It globally minimizes the the length $\cL_\cF$ with constraint that $\cG_{i,j}(0)\in\cS_i$ and $\cG_{i,j}(1)\in\cS_j$. In the remainder of this paper, We referred to  $\cG_{i,j}$ as the \emph{connection path} between a pair of adjacent boundary proposals $\cS_i$ and $\cS_j$.

The geodesic distance between the boundary proposals $\cS_i$ and $\cS_j$,  denoted by  $\Dist_\cF$, is defined as
\begin{equation}
\Dist_\cF(\cS_i,\cS_j)=\min_{\gamma\in\Lip([0,1],\bM)}\cL_\cF(\gamma), ~s.t.
\begin{cases}
\gamma(0)\in\cS_i&\\
\gamma(1)\in\cS_j&	
\end{cases}
\end{equation}
This defines a geodesic distance map $\cD_{\cS_i}$ associated to a fixed $\cS_i$ such that
\begin{equation*}
\cD_{\cS_i}(\fx)=\Dist_\cF(\cS_i,\cS_j).
\end{equation*}
The geodesic distance map $\cD_{\cS_i}$ can be efficiently solved by the Hamilton Fast Marching method\footnote{The code for the HFM method involving the computation of the geodesic distances  and the tracking of the geodesic paths can be downloaded from~\href{https://github.com/Mirebeau/HamiltonFastMarching}{https://github.com/Mirebeau/HamiltonFastMarching}.} (HFM)~\cite{mirebeau2018fast,mirebeau2019riemannian,mirebeau2021massively}.

Finally, we estimate a solution, denoted by $\cG$, to a generalized gradient descent ODE
\begin{equation}
\label{eq:GeoPathODE}
\cG^\prime(u)=\max_{\|\fu\|=1}\frac{\langle\nabla\cD_{\cS_i}(\cG(u)),\fu\rangle}{\cF(\cG(u),\fu)}~\text{with}~\cG(0)=\fy^*_{i,j}.
\end{equation}
The connection path $\cG_{i,j}$ can be obtained by re-parameterizing solution $\cG$, such that 
\begin{equation}
\label{eq_ConnectPoints}
\begin{cases}
\cG_{i,j}(0)=\fx^*_{i,j}\in\cS_i	\\
\cG_{i,j}(1)=\fy^*_{i,j}\in\cS_j.
\end{cases}	
\end{equation}
The points $\fx^*_{i,j}$ and  $\fy^*_{i,j}$ play a crucial role in building the final segmentation contour, as discussed in Section~\ref{sub_CircularShortestPath}.

\subsubsection{Computing the Edge Weights}
In the context of image segmentation, the edge weight  $w_{i,j}$ should be low if both of the boundary proposals $\cS_i$ and $\cS_j$ are a part of the target boundary. In our model, the computation of  the weight $w_{i,j}$  leverages an energy $\kC:\Lip([0,1],\Omega)\to\bR^+$ measured along the connection path $\cG_{i,j}$ such that $w_{i,j}=\infty$ if the node $\vartheta_i$ are \emph{not} adjacent to $\vartheta_j$, and otherwise 
\begin{equation}
\label{eq:weightFunc}
w_{i,j}:=
\begin{cases}
\kC(\cG_{i,j}),&\forall u\in[0,1],~\cG_{i,j}(u)\notin\rC_{\fp},\\
\infty,&\exists u\in[0,1],~\cG_{i,j}(u)\in\rC_{\fp}.
\end{cases}
\end{equation}

In this paper, we provide two methods to formulating the energy $\kC$, among which  the energy $\kC(\cG_{i,j})$ of the connection path $\cG_{i,j}$ features a common part that is the euclidean length of $\cG_{i,j}$. Specifically, the first choice for $\kC:=\kC_1$ is defined as
\begin{equation}
\label{eq_GeoEnergy1}
\kC_1(\cG_{i,j})=\mu\Length(\cG_{i,j})+\cL_{\cF}(\cG_{i,j}),
\end{equation}
where $\Length(\cG_{i,j})$ is the euclidean length of $\cG_{i,j}$, and $\mu>0$ is a constant parameter.

The second choice for the energy $\kC:=\kC_2$ relies on the curvature $\kappa_{i,j}:[0,1]\to\bR$ of $\cG_{i,j}$, i.e.
\begin{equation}
\label{eq_CurvaEnergy}
\kC_2(\cG_{i,j})=\int_0^1 \sqrt{1+\beta^2\kappa_{i,j}(u)^2}\|\cG_{i,j}^\prime(u)\|du,
\end{equation}
where $\beta>0$ is a constant as a weighting parameter and $\cG_{i,j}^\prime$ is the first-order derivative of the connection path $\cG_{i,j}$.

By the definition~\eqref{eq:weightFunc}, we set the weight $\omega_{i,j}=+\infty$ if the connection paths $\cG_{i,j}$ or $\cG_{j,i}$ intersect with the adaptive cut. This leads to the fact that the nodes $\vartheta_i$ and $\vartheta_j$  corresponding to the boundary proposals $\cS_i$ and $\cS_j$ are \emph{not} adjacent in the graph. As a result, the use of the adaptive cut imposes disconnection on the graph, providing the possibility of finding circular paths from this graph. Furthermore, we take into account a direct graph for computing shortest paths, i.e. we allow the weights obeying $\cG_{i,j}\neq\cG_{j,i}$.

\begin{figure*}[t]
\centering{
\includegraphics[width=0.98\linewidth]{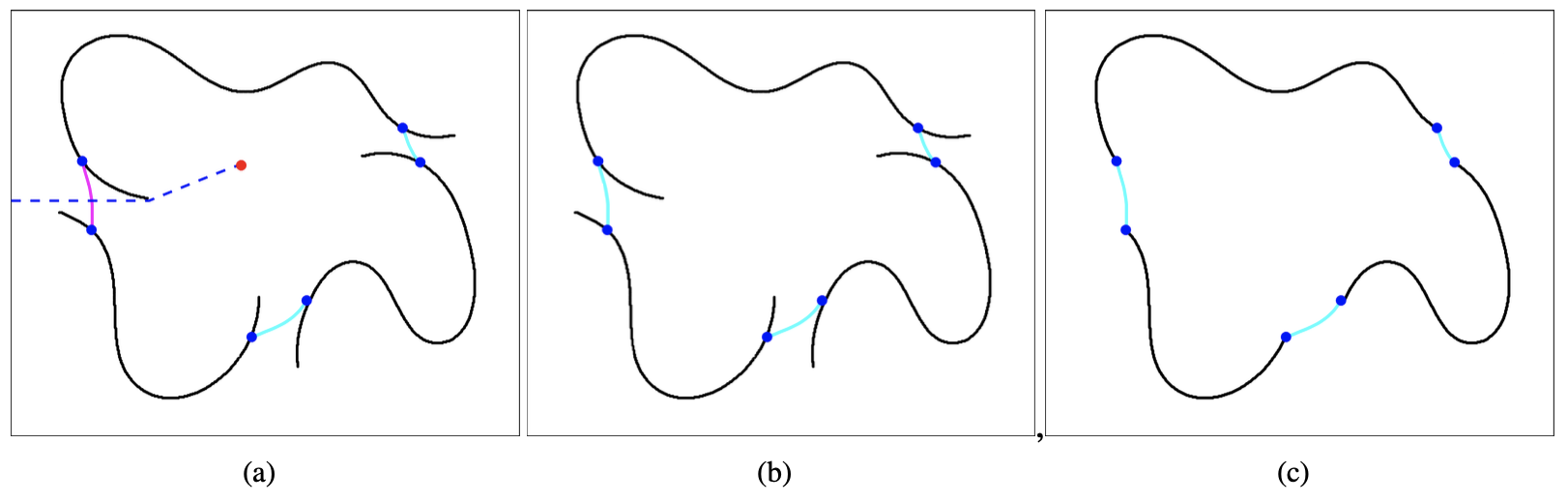}
}
\caption{An example for building the truncated boundary proposals. (\textbf{a}) The black lines indicate the detected disjoint boundary proposals. The red dot is the landmark point. The blue dashed line denotes the adaptive cut. The cyan and magenta lines are the connection paths. Note that the magenta line intersects with the adaptive cut. The blue dots represent the intersecting points between the boundary proposals and the connection paths. (\textbf{b}) The boundary proposals and the connection paths (cyan lines).  (\textbf{c}) The truncated boundary proposals and the corresponding connection paths which form a simple closed contour.}
\label{fig:gapsExamples}	
\end{figure*}

\subsection{Tracking Circular Optimal Paths}
\label{sub_CircularShortestPath}
A discrete path $\Gamma$ in the discrete graph $G=(\cV,\cE)$ is made up of a series of nodes. The length of a path $\Gamma$ is defined as
\begin{equation}
\label{eq_EnergyDiscrete}
\kL(\Gamma)=\sum_{i,j}w_{i,j},	
\end{equation}
where  $w_{i,j}$ is the weight for the edge $e_{i,j}\in\cE$ such that its corresponding nodes $\vartheta_i$ and $\vartheta_{j}$ are successive points in the discrete path $\Gamma$. The  Dijkstra's algorithm~\cite{dijkstra1959note} is known as an efficient way to extract an open shortest path from a discrete graph $G$.  Given two nodes $\vartheta_i,\,\vartheta_j\in\cV$, one can extract a discrete shortest path $\tilde\cP_{i,j}$ via Dijkstra's algorithm that links from $\vartheta_i$ to $\vartheta_j$. 
In principle, the nodes $\vartheta_i$ and $\vartheta_j$ are not connected in the sense of the discrete path $\tilde\cP_{i,j}$. In order to obtain a closed continuous contour for segmentation, the nodes $\vartheta_i$ and $\vartheta_j$ should be connected in an appropriate way. In this section, we use the adaptive cut $\rC_\fp$ to achieve this purpose. 

We build a set $\Lambda$ involving all the pairs of ordered nodes $(\vartheta_i,\vartheta_j)$ whose  connection path intersects with the adaptive cut $\rC_{\fp}$, see Fig.~\ref{fig:gapsExamples}a for a typical example of  such a pair of nodes. For each  pair of nodes $(\vartheta_i,\vartheta_j)\in\Lambda$,  we can extract a discrete shortest path $\cP_{i\to j}$ by Dijkstra's algorithm, which links from the source node $\vartheta_i$ to the target node $\vartheta_j$. 

Each node $\vartheta_i$ in the discrete shortest path $\cP$ corresponds to a boundary proposal $\cS_i$ in the image domain. In other words, $\cP$ determines a set of successive boundary proposals, as illustrated in Fig.~\ref{fig:gapsExamples}b.  Each boundary proposal $\cS_i$ involved in $\cP_{i\to j}$ has two connection paths, and each path  intersects with $\cS_i$ at two points, see Eq.~\eqref{eq_ConnectPoints}. We denote by $\tilde\cS_i\subset \cS_i$ the truncated segment of $\cS_i$ between those intersection points. Eventually, one can simply identify a circular optimal path $\cC_{i\to j}$, as the concatenation of the truncated boundary proposals involved in $\cP_{i\to j}$ and the corresponding connection paths. As a result, using all pairs of nodes in the set $\Lambda$, we can generate a set of circular optimal paths, each of which encloses the landmark point $\fp$ and intersects with the adaptive cut $\rC_\fp$ only once. All  circular optimal paths form an \emph{admissible} set 
\begin{equation}
\chi_\fp:=\{\cC_{i\to j}\,|\,\forall (\vartheta_i,\vartheta_j)\in\Lambda\}.
\end{equation}
We now  search for an optimal contour from the set $\chi_\fp$ by minimizing the following energy functional 
\begin{equation}
\label{eq:OptimalCircularPath}
\cC^*=\underset{\cC\in{\chi_\fp}}{\arg\min}\, \left\{\mu_1\kL(\cP)+\mu_2\Length(\cC)^{-1}\right\},
\end{equation}
where the parameters $\mu_1$, $\mu_2$ are positive constants. The term $\Length(\cC)$ is the euclidean curve length of the circular path $\cC\in\chi_\fp$. The term $\kL(\cP)$  is the summation of the edge weights, as formulated in Eq.~\eqref{eq_EnergyDiscrete}. By Eq.~\eqref{eq:OptimalCircularPath}, the optimal contour is expected to have high euclidean curve length to avoid small contours enclosing the point $\fp$ and low connection cost to encourage $\cC^*$ to admit the target boundary.

\subsection{Discussion on the Computation Complexity}
In the proposed method, the computation cost mainly contains two parts.
The first part is the construction of the discrete graph, involving the detection of the adjacent boundary proposals, the computation of connection paths, and the estimation of the graph edge weights. Indeed, this step is very time-consuming especially for the estimation of the edge weights. However, the computation of this step can be greatly speeded up by a parallel computing implementation. In addition, the graph construction does not rely on user intervention, thus can be carried \emph{offline}.
The second part is the interactive segmentation step, involving the computation of the adaptive cut and the construction of the optimal circular paths. Specifically, the computation of the adaptive cut is implemented by the classical fast marching method, with a complexity $\mathcal{O}(N\log N)$, where $N$ represents the number of grid points in the discrete domain $h\bZ^2\cap\Omega$ with $h$ being the grid scale. The complexity of the establishment of the set $\chi_{\fp}$ is $\mathcal{O}(mM\log M)$, where $m$ is the number of elements in the set $\Lambda$ and $M$ is the number of the boundary proposals. As a result, the interactive segmentation step of the proposed model can achieve a real-time manner, thanks to the sparsity of the boundary proposals.

\begin{figure*}[t]
\centering{
\includegraphics[width=0.98\linewidth]{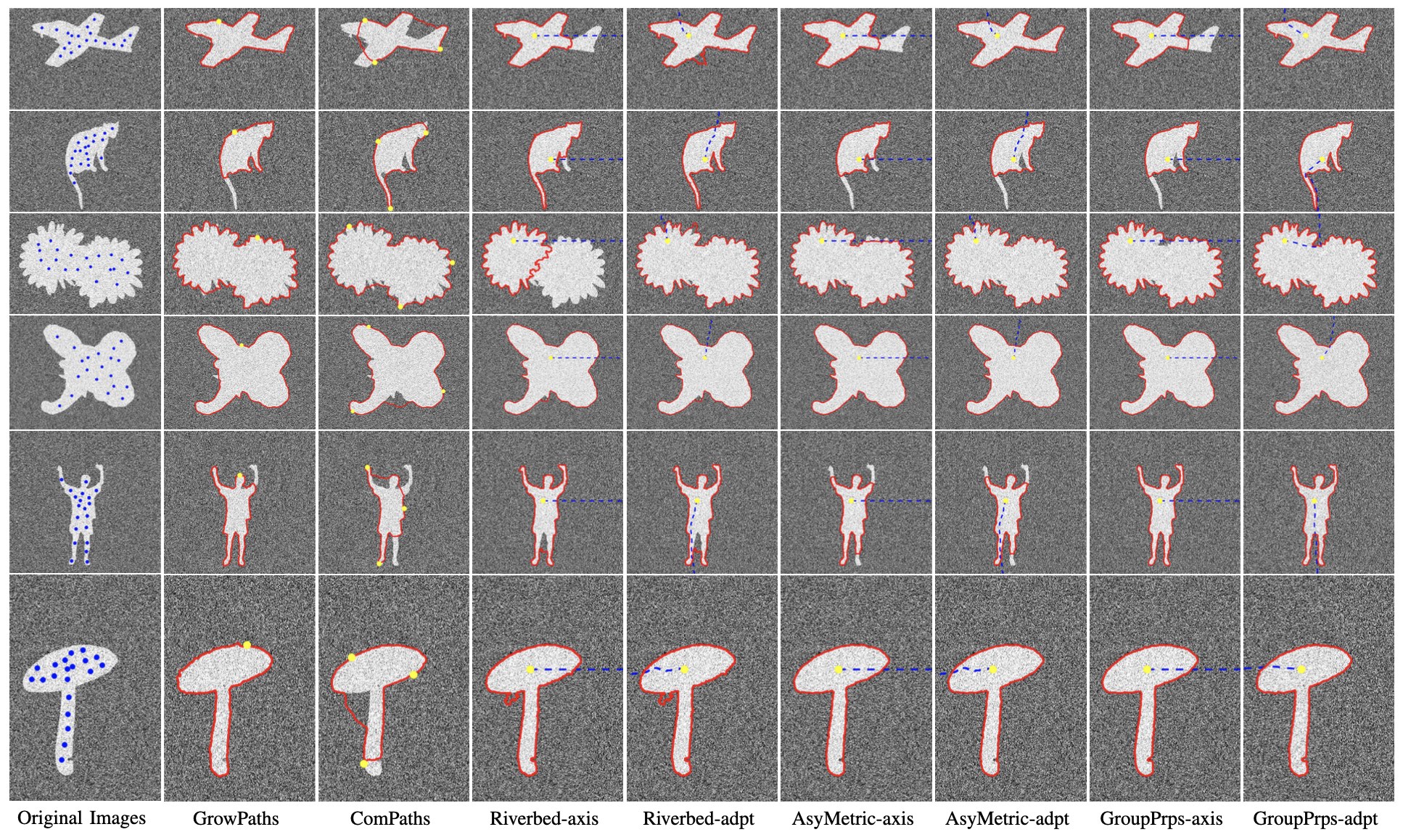}
}
\caption{Segmentation results on synthetic images interrupted by additive Gaussian noise with normalized standard derivation $\sigma_n=0.125$. The red lines represent the segmentation contours and the landmark points are denoted by yellow dots. The blue dash lines stand for the cuts.  \textbf{Column 1}: The original synthetic images  and  $20$ landmark points indicated by blue dots. \textbf{Columns 2-9}: The closed contours extracted from the GrowPaths model~\cite{benmansour2009fast}, the ComPaths model~\cite{mille2015combination}, the Riverbed-axis model, the Riverbed-adapt model, the AsyMetric-axis model, the AsyMetric-adpt model, the GroupPrps-axis model and the GroupPrps-adapt model, respectively}
\label{fig:syn}
\end{figure*}

\begin{figure}[t]
\centering
\includegraphics[width=8cm]{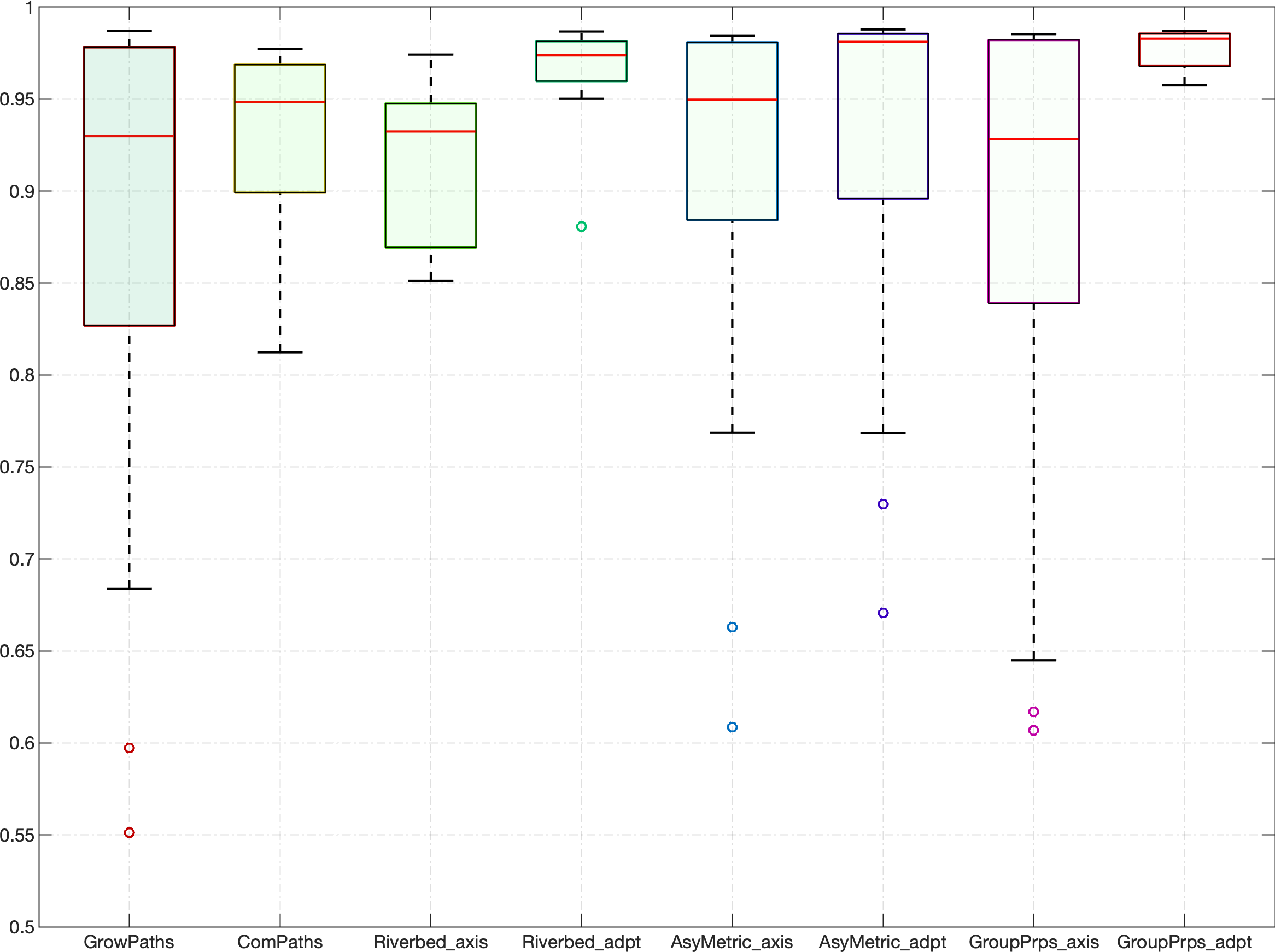}
\caption{Box plots of the average Dice scores $\bD$ of 20 runs per image over all synthetic images for different methods.}
\label{fig:examples_syn}	
\end{figure}

\begin{figure}[t]
\centering
\includegraphics[width=8cm]{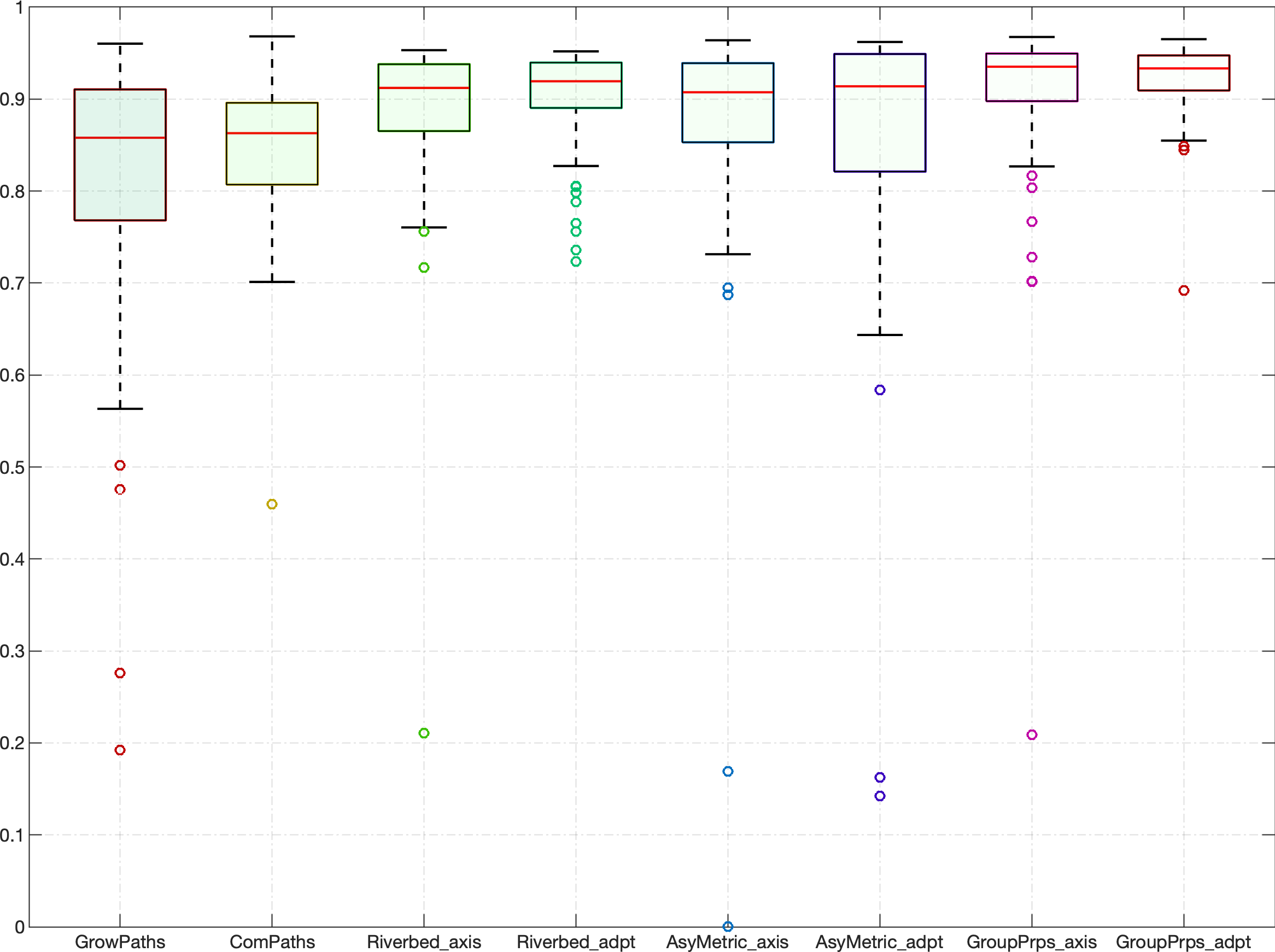}
\caption{Box plots of the average Dice scores $\bD$ of 10 runs per image over CT dataset for different methods.}
\label{fig:examples_CT}	
\end{figure}

\renewcommand{\arraystretch}{1.5}
\begin{table*}[t]
  \centering
  \fontsize{8}{10}\selectfont
  \begin{threeparttable}
  \caption{The quantitative comparison results of the GrowPaths, ComPaths, Riverbed-axis, Riverbed-adapt,  AsyMetric-axis, AsyMetric-adpt, GroupPrps-axis and GroupPrps-adapt models on synthetic images.}
  \label{tab:performance_comparison_syntheticImages}
  \setlength{\tabcolsep}{3pt}
\renewcommand{\arraystretch}{1.5}
    \begin{tabular}{ccccccccccccccccccccc}
        \toprule
    \multirow{2}{*}{Noise Level}&
    \multicolumn{2}{c}{GrowPaths}&\multicolumn{2}{c}{ComPaths}&\multicolumn{2}{c}{Riverbed-axis}&\multicolumn{2}{c}{Riverbed-adpt}&\multicolumn{2}{c}{AsyMetric-axis}&\multicolumn{2}{c}{AsyMetric-adpt}&\multicolumn{2}{c}{GroupPrps-axis}&\multicolumn{2}{c}{GroupPrps-adpt}\cr
    \cmidrule(lr){2-3} \cmidrule(lr){4-5} \cmidrule(lr){6-7} \cmidrule(lr){8-9} \cmidrule(lr){10-11}\cmidrule(lr){12-13}\cmidrule(lr){14-15}\cmidrule(lr){16-17}
    &Mean&Std&Mean&Std&Mean&Std&Mean&Std&Mean&Std&Mean&Std&Mean&Std&Mean&Std\cr
    \midrule
    0.025& 0.9137&0.2105 &0.9480&0.0544&0.9156 &0.1835 &0.9709&0.0112&0.9424& 0.1551& 0.9619&0.1191 & 0.8949 & 0.2537&{\bf0.9793}&{\bf0.0084}\cr
    0.050& 0.8832&0.2480 &0.9394&0.0582&0.9273 &0.1700 &0.9722&0.0108& 0.9225&0.1906 &0.9432&0.1588&0.8916&  0.2557&{\bf0.9792}&{\bf0.0084}\cr
    0.075& 0.9257&0.1764 &0.9253&0.0716&0.9200 & 0.1634&0.9728&0.0106&0.9132& 0.1987&0.9334&0.1771& 0.8770&0.2571&{\bf0.9790}&{\bf0.0083}\cr
    0.100& 0.8759&0.2379 &0.9254&0.0597&0.9235 &0.1743&0.9731&0.0108&0.9011&  0.2141&0.9081&0.2219& 0.8770 &0.2418&{\bf0.9782}&{\bf0.0084}\cr
    0.125& 0.8114&0.2597 &0.9045&0.0702&0.9152&0.1776 &0.9541&0.1046&0.8852& 0.2352&0.9252&0.1785&0.8766& 0.2329&{\bf0.9754}&{\bf0.0120}\cr
    \bottomrule
    \end{tabular}
    \end{threeparttable}
\end{table*}

\renewcommand{\arraystretch}{1.5}
\begin{table*}[]
  \centering
  \fontsize{8}{10}\selectfont
  \begin{threeparttable}
  \caption{The quantitative comparison results of the GrowPaths, ComPaths, Riverbed-axis, Riverbed-adapt,  AsyMetric-axis, AsyMetric-adpt, GroupPrps-axis and GroupPrps-adapt models on CT images.}
  \label{tab:performance_comparison_realImage}
    \setlength{\tabcolsep}{3pt}
\renewcommand{\arraystretch}{1.5}
    \begin{tabular}{ccccccccccccccccccccc}
    \toprule

    \multicolumn{2}{c}{GrowPaths}&\multicolumn{2}{c}{ComPaths}&\multicolumn{2}{c}{Riverbed-axis}&\multicolumn{2}{c}{Riverbed-adpt}&\multicolumn{2}{c}{AsyMetric-axis}&\multicolumn{2}{c}{AsyMetric-adpt}&\multicolumn{2}{c}{GroupPrps-axis}&\multicolumn{2}{c}{GroupPrps-adpt}\cr\cmidrule(lr){1-2}
    \cmidrule(lr){3-4} \cmidrule(lr){5-6} \cmidrule(lr){7-8} \cmidrule(lr){9-10} \cmidrule(lr){11-12}\cmidrule(lr){13-14}\cmidrule(lr){15-16}\cmidrule(lr){17-18}
    Mean&Std&Mean&Std&Mean&Std&Mean&Std&Mean&Std&Mean&Std&Mean&Std&Mean&Std\cr

    \midrule
     0.8184&0.1971 &0.8529&0.1277 &0.8859&0.1335&0.9037& 0.0850&0.8733&0.1488&0.8683&0.1527&0.9063& 0.1081&{\bf0.9241}&{\bf0.0718}\cr
    \bottomrule
    \end{tabular}
    \end{threeparttable}
\end{table*}

\renewcommand{\arraystretch}{1.5}
\begin{table*}[]
  \centering
  \fontsize{8}{10}\selectfont
  \begin{threeparttable}
  \caption{The execution time of the GrowPaths, ComPaths,  Riverbed-adapt, AsyMetric-adpt, and GroupPrps-adapt models on synthetic and real images.}
  \label{tab:executionTime}
    \setlength{\tabcolsep}{11pt}
\renewcommand{\arraystretch}{1.5}
    \begin{tabular}{ccccc}
    \toprule  
    GrowPaths&ComPaths&Riverbed-adpt&AsyMetric-adpt&GroupPrps-adapt\cr
    \midrule
     0.8184&0.1971 &0.8529&0.1277&{\bf0.0718}\cr
    \bottomrule
    \end{tabular}
    \end{threeparttable}
\end{table*}

\begin{figure*}[t]
\centering{
\includegraphics[width=0.98\linewidth]{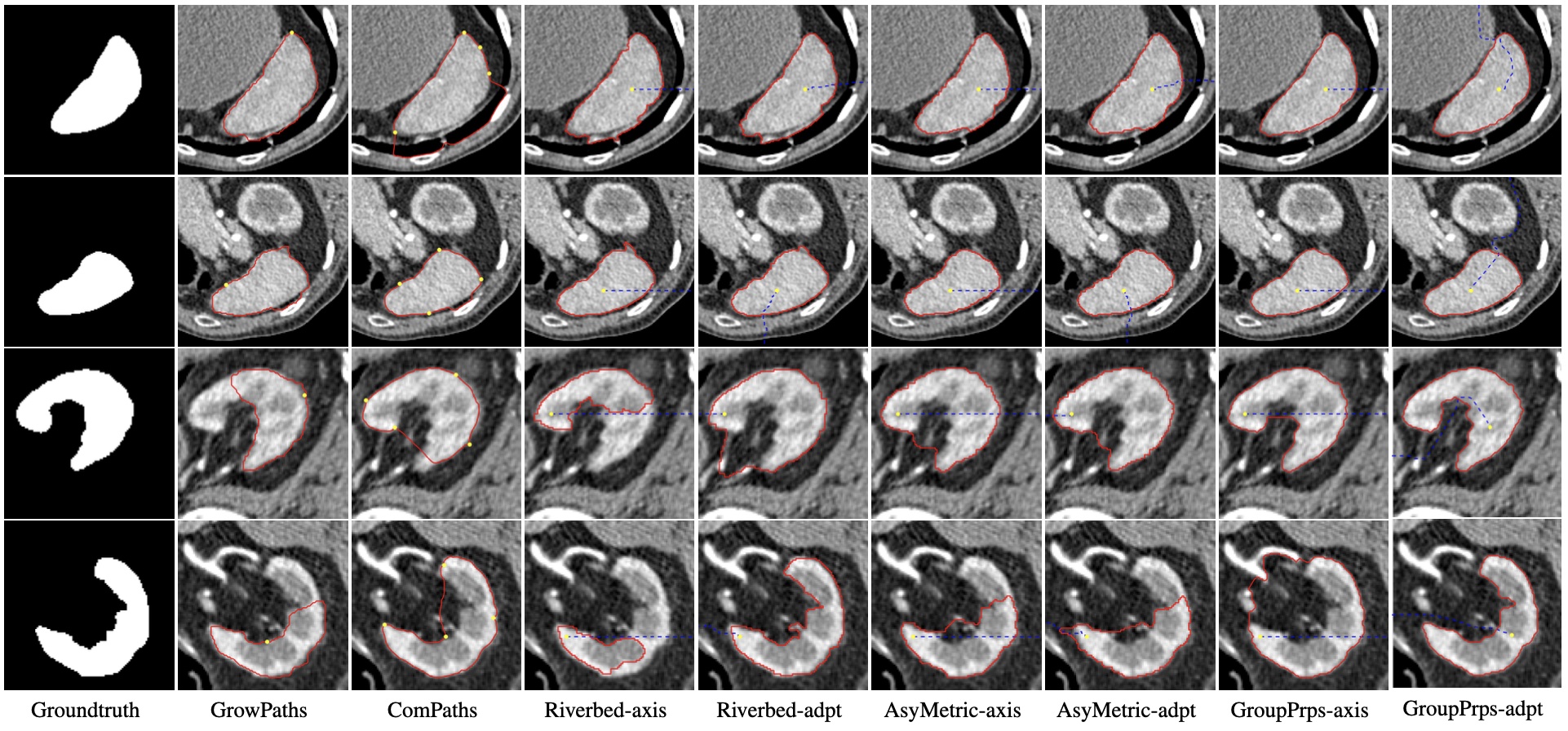}}
\caption{Qualitative comparison results from different models on CT images. The yellow dots denote the user-provided landmark points and the red lines indicate the obtained closed paths. The cuts are denoted by blue dash lines. \textbf{Column} 1: Ground Truth. \textbf{Columns 2-9} : Results from the GrowPaths model, the ComPaths model, the Riverbed-axis model, the Riverbed-adapt model, the AsyMetric-axis model, the AsyMetric-adpt model the GroupPrps-axis model and the GroupPrps-adapt model, respectively}
\label{fig:CTImg}
\end{figure*}

\begin{figure*}[t]
\centering{
\includegraphics[width=0.98\linewidth]{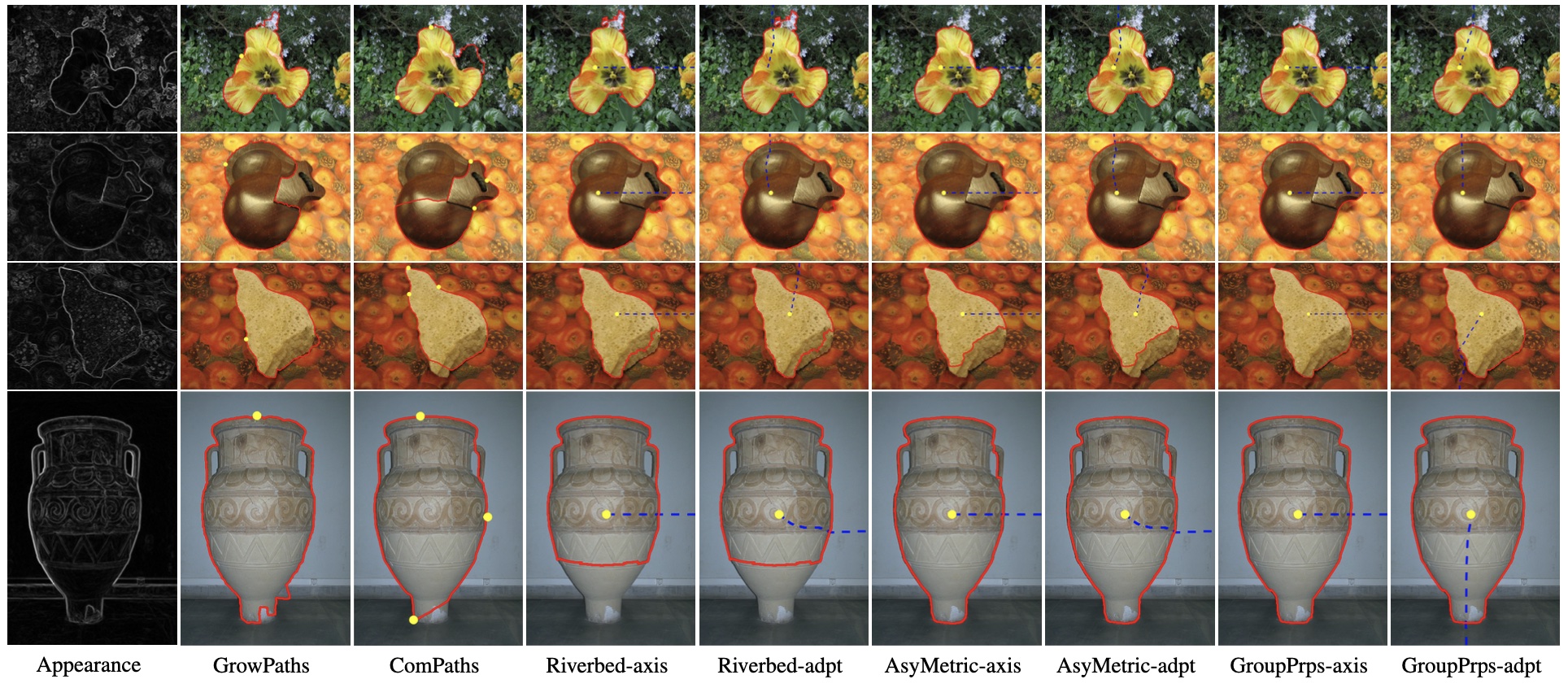}
}
\caption{Qualitative comparison results from different models on nature images. The yellow dots indicate the user-provided landmark points and the red lines represent the obtained closed paths. The axis cuts and adaptive cuts are denoted by blue dash lines \textbf{Column} 1: Visualization for the edge-based appearance features. \textbf{Columns} 2-9: Results from the GrowPaths model, the ComPaths model, the Riverbed-axis model, the Riverbed-adapt model, the AsyMetric-axis model, the AsyMetric-adpt model, the GroupPrps-axis model and the GroupPrps-adapt model, respectively}
\label{fig:natureImg}
\end{figure*}

\begin{figure*}[t]
\centering{
\includegraphics[width=0.98\linewidth]{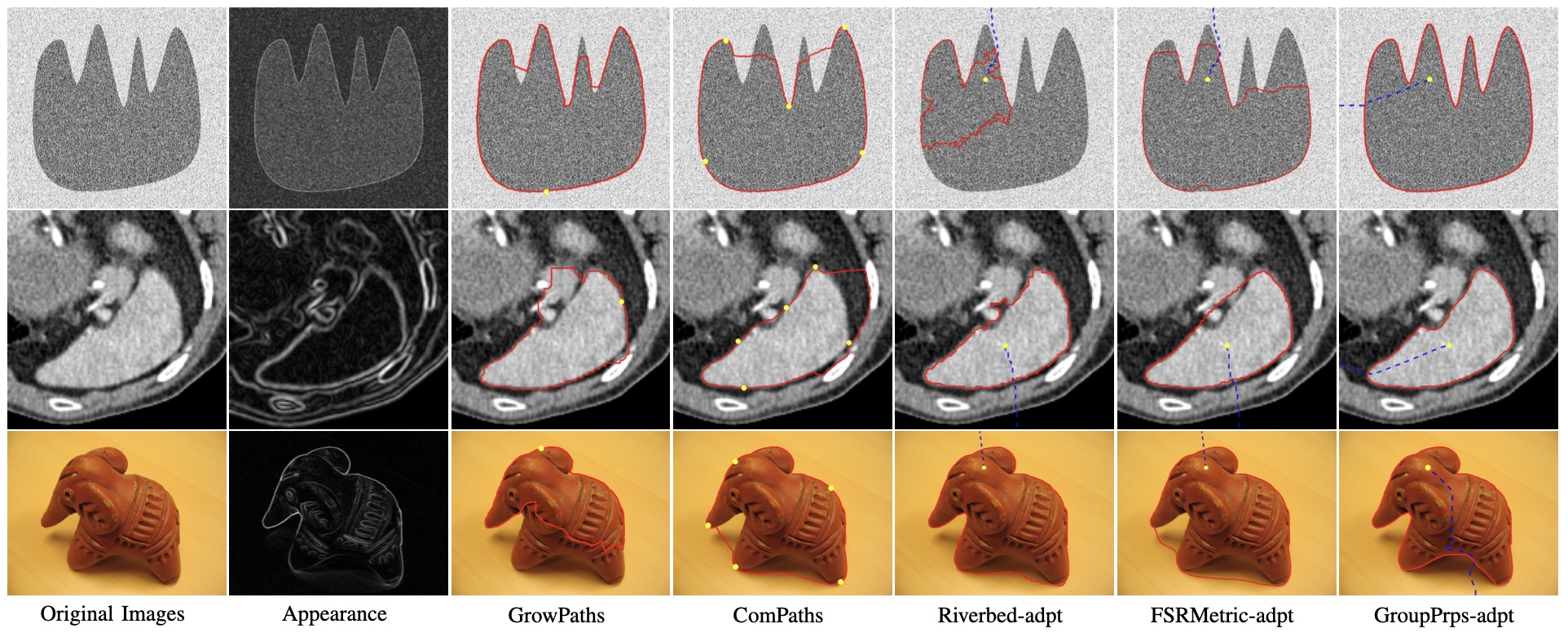}}
\caption{Qualitative comparison results from different models. The yellow dots indicate the user-provided landmark points and the red lines represent the obtained closed paths. The cuts are denoted by blue dash lines. \textbf{Column}  1: Original images. \textbf{Column} 2: Visualization for the edge-based appearance features. \textbf{Columns} 2-6: Results from the GrowPaths, ComPaths, Riverbed-adpt, FSRMetric-adpt and the proposed GroupPrps-adpt models, respectively}
\label{fig:CurvatureEnergy}
\end{figure*}

 \begin{figure*}[t]
\centering
\includegraphics[width=0.98\linewidth]{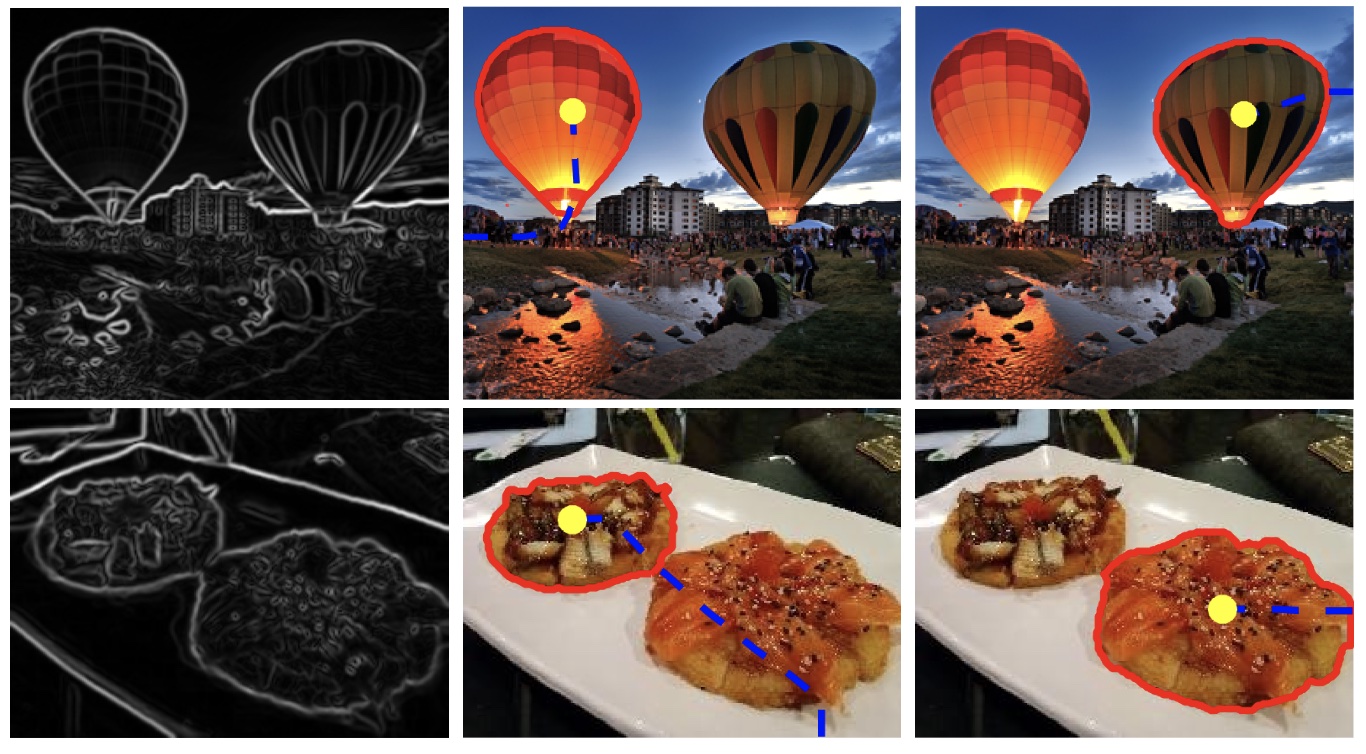}
\caption{Examples for multiple object detection by the proposed GroupPrps model. \textbf{Column} 1: Visualization for the edge-based appearance features. \textbf{Columns 2} and \textbf{3}: Results from the GroupPrps model. Cyan or red lines indicate different object structures. The blue dash lines depicts the adaptive cuts. The yellow dots represent the initial points}
\label{fig:multiObjects}	
\end{figure*}

\section{Experimental Results} 
\label{sec:Exp}

In this section, the numerical experiments are conducted to compare the proposed model (referred to as GroupPrps) with the growing minimal paths (GrowPaths) model~\cite{benmansour2009fast}, the combination of piecewise-geodesic paths (ComPaths) model~\cite{mille2015combination}, the user-steered optimum boundary tracking model (Riverbed)~\cite{miranda2012riverbed} and the dual-cut model~\cite{chen2021geodesic} with a spatial asymmetric quadratic metric (AsyMetric) or a reeds-shepp forward metric (FSRMetric). The Dice index $\bD$ is adopted for the quantitative comparisons to evaluate the segmentation results defined as follows:
\begin{equation}
\label{eq:accuracyscore}
\bD(S,GT)=\frac{2|S\cap GT|}{|S|+|GT|},
\end{equation}
where $S$ stands for the segmented region, $GT$ presents the ground truth, and $|S|$ denotes the area of $S$. The Dice index $\bD$ is ranged from 0 to 1, when $\bD=1$ means that the segmented result is completely overlapped with its corresponding ground truth region. 

\subsection{Parameter Configuration}
\label{subsec:Parameter}

In the following experiments, the edge indicator $\phi$ defined in Eq.~\eqref{eq:EdgeDetector} is computed by setting the parameters $\tau=1$ and $w=0.1$. In the graph construction step, the width of tubular neighbourhood of the boundary proposal is set as $12$ grid points, see Eq.~\eqref{eq:Neighbourhood}. For recovering the circular optimal path, the parameters $\mu_1$ and $\mu_2$ are used to control the relative importance of the path cost and path length, respectively in Eq.~\eqref{eq:OptimalCircularPath}. We set $\mu_1=1$ and $\mu_2=0.1$ in the experiments. The input of the proposed model is a landmark point inside the target region. The GrowPaths model computes a set of geodesic paths in conjunction with a sequentially keypoints detection scheme during the front propagation. Thus this model is able to achieve a fast object segmentation from a single input along the target boundary. In the following experiments, we set up the GrowPaths model such that the scale of two keypoints is fixed as $10$ grid points. The ComPaths model extracts a closed contour by connecting a set of user-provided points using piecewise-geodesic paths. The maximal number of admissible paths for each pair of successive vertices is chosen as 5 for all the tests. For the Riverbed, AsyMetric and GroupPrps models, we take advantages of the strategy of providing the landmark point within the target region to start the segmentation procedure. The non-negative $x$-axis cut or the proposed adaptive is computed from the landmark point to select the initial point of the closed target boundary. In the Riverbed model, the initial point is the intersection point of the cut with the object boundary generated from its ground truth. The Riverbed models with the x-axis cut and the adaptive cut is regarded as Riverbed-axis and Riverbed-adpt models, respectively. For the AsyMetric model, the initial point is detected automatically by finding the point with maximal  image gradient magnitude value along the cut. The AsyMetric models with the $x$-axis cut and the adaptive cut is regarded as AsyMetric-axis and AsyMetric-adpt models, respectively. In the experiments, we take into account the Euclidean path length-based energy to estimate weight for the edge to connect adjacent nodes in the graph construction. The energy is computed simultaneously during the fast propagation based on the isotropic Riemannian metric. The proposed GroupPrps models with the $x$-axis cut and the adaptive cut is regarded as GroupPrps-axis and GroupPrps-adpt models, respectively. 

Finally, all the experiments are performed on a standard $8$-core Intel Core i7 of $3.8$GHz architecture with $64$Gb RAM.

\subsection{Quantitative and Qualitative Comparison Results}
\label{subsec:Comparison}

We conduct experiments on synthetic images, CT images~\cite{spencer2018parameter} and some nature images from the GrabCut dataset~\cite{rother2004grabcut} to compare the segmentation models quantitatively and qualitatively in this section. The synthetic images those are constructed based on the ground truth from the dataset~\cite{rother2004grabcut}. The intensity values of all pixels corresponding to the target structures and the background region on the ground truth are fixed as 1 and 0.5, respectively. Each image is interrupted by additive Gaussian noise whose normalized standard derivation values are set as $5$ levels (from $\sigma_n=0.025$ to $0.125$), yielding  $30$ synthetic images for the following experiments. The used $6$ synthetic images with noise level $\sigma_n=0.025$ are exhibited in column $1$ of Fig.~\ref{fig:syn}. 

We firstly conduct the experiment on the synthetic images to compare the robustness of different models with respect to different initializations. For the Riverbed, AsyMetric and GroupPrps models, we choose $20$ landmark points inside the target region of each synthetic image as the inputs.  The selected $20$ landmark points are evenly distributed inside the target region, as illustrated by blue points in column $1$ of Fig.~\ref{fig:syn}.  Then we compute an adaptive cut from each landmark point. Furthermore, the  axis cut  along the non-negative $x$-axis is also computed. 
Therefore the effectiveness of the proposed adaptive cut is analyzed in the experiments. For the GrowPaths model, $20$ seed points are sampled along the boundary of the target regions for initializing the model. For the ComPaths model, we build $20$ groups of sampled points for initialization, each of which involves three points extracted in a clockwise order along the boundary of the target structure. With these initializations, all the considered models are conducted $20$ runs per image. The average of the Dice segmentation accuracy index over the $20$ runs for each image is computed from different models. As in Fig.~\ref{fig:examples_syn}, we observe that similar segmentation results are obtained with respect to different landmark points, while the other compared models are sensitive to the positions of the landmark points.

The experiment on synthetic images blurred by noise of different levels is conducted to further demonstrate the robustness of the proposed method to noise. All the considered models are conducted $20$ runs with respect to each initialization per image. The average and standard deviation values of the Dice index under different levels of noises are illustrated in Table~\ref{tab:performance_comparison_syntheticImages}.  It shows that the Dice index values derived from the proposed model are slightly affected by different kinds of noise with different levels, and have the best performance among all the compared models. We can see that the proposed GroupPrps-adpt model achieves the best robustness performance against to noise, benefitting from the boundary proposal prior and the global optimality of the graph-based path searching scheme. In addition, the effeteness of the adaptive cut can be demonstrated as well. Comparing the Dice index values from the axis cut-based models with those from the adaptive cut-based models, the use of the adaptive cut can improve the segmentation accuracy significantly. 

In Fig.~\ref{fig:syn}, we present the qualitative performance of the compared models on the synthetic images interrupted by additive Gaussian noise with normalized standard derivation $\sigma_n=0.125$ . The segmented results from the GrowPaths, ComPaths, Riverbed-axis, Riverbed-adapt, AsyMetric-axis, AsyMetric-adpt, GroupPrps-axis and GroupPrps-adapt models are depicted in Columns 2 and 9 of Fig.~\ref{fig:syn}. The cut from the internal point is depicted as the blue dashed line. The extraction contours are described by red lines and the landmark points are depicted as yellow dots. One can get that the GrowPaths model is prone to missing small parts of the target structure and the ComPaths model might be stuck in unexpected positions. In the top three rows, the  axis cut has more than one intersection points with the object boundary. The Riverbed-axis, AsyMetric-axis and GroupPrps-axis models fail to extract the correct boundaries due to that the detected contour can pass through the cut only one time. In the AsyMetric model~\cite{chen2021geodesic}, a complex dual-cut strategy is utilized to solve this problem. In the proposed model, the adaptive cut has exactly one intersection point with the boundary. Therefore, the adaptive cut scheme improves the segmentation accuracy effectively. In rows $4$ to $6$, the axis cut and  the adaptive cut both cross with the boundary one time. They have similar detected results. Finally, we get that the region of interest is successfully extracted as the closed contour tracked from the adaptive cut by the proposed model.

Furthermore, we perform the compared experiments quantitatively and qualitatively of different models on $86$ CT images. Each image is artificially interrupted by additive Gaussian noise of normalized standard derivation $\sigma_n=0.025$. We also test the robustness of different models to initializations on CT images. We apply the same way of providing initializations with different positions for each model as described above. More specifically, for each model, $10$ landmark points are generated inside the target region per image. Each landmark point is regarded as the initial point that the cut emanated from. In addition, $10$ seed points along the target boundary of each image are specified to set up the GrowPaths model. In this case, we detect 10 groups of four landmark points in a clockwise order along the boundary of the target region per image for the ComPaths model.  

The quantitative comparison results of 10 runs per image over a set of $86$ CT images are carried out by means of Dice index. The average Dice accuracy index of $10$ runs for each image over the CT set is illustrated in Fig.~\ref{fig:examples_CT}. The experiments on CT images also show that the proposed model is robust to the initialization positions. Besides, the average and standard deviation values of the Dice index of all comparable models are listed in Table~\ref{tab:performance_comparison_realImage}. We get that the proposed GroupPrps-adapt model achieve best accuracy performance among all the comparable models. The qualitative comparison results of the considered models are illustrated in Fig.~\ref{fig:CTImg}. The target structures in the presence of intensity inhomogeneity are surrounded by complicated background. There also exist parts of blurred boundaries with low gray levels. In Fig.~\ref{fig:CTImg}, we represent the detected closed contours by red lines on CT images. Columns $2$ and $3$ illustrate the results from the GrowPaths and ComPaths models, respectively. One can observe that segmentation problems may occur along the blurred edges. From columns $4$ to $9$, we get that the adaptive cut pass through the object boundary only once which guarantees the target contour is a simple closed contour along the landmark point. The boundary proposal grouping scheme can track the  boundary of the target structure successfully from the adaptive cut.

In Fig.~\ref{fig:natureImg}, we qualitatively compare different models mentioned above on nature images. Our goal is to delineate the target boundary by detecting the closed contour with user-provided points. The target structures with varied shapes are located in complicated background or have blurred edges with the background. The segmented results are depicted as red lines and the generated cut is shown as blue dash line. The landmark points are illustrated as yellow dots. In column $1$, we exhibit the edge-based appearance features, by which the isotropic geodesic metric is constructed. The detected results of the GrowPaths model are depicted in column 2. We can observe that it is not able to track the correct boundary where the edge-based appearance features are weak. The extracted results by the ComPaths model with randomly-located initial seed points are demonstrated in column $3$. The ComPaths model initialized with unevenly spaced seed points might cause undesirable shortcut problem. The final combination paths fail to describe the target boundaries correctly. From columns $4$ to $9$, we illustrate the axis cut, the adaptive cut and the detected results of Riverbed, AsyMetric and GroupPrps models. The proposed model can extract the target structures correctly.

We compare the execution times of the rowPaths, ComPaths,  Riverbed-adapt, AsyMetric-adpt, and GroupPrps-adapt models on synthetic and real images illustrated in the first column of Fig.~\ref{fig:CurvatureEnergy}. The mean size of the three images is $435 \times 470$ grid points. In the proposed model, only the execution time of interactive segmentation step  including the adaptive cut computation and the optimal circular path construction is considering, since the graph construction step can be carried offline. The results of execution times are shown in Table~\ref{tab:executionTime}. One can get that the proposed method has lower execution time. The target contour can be extracted immediately after the landmark point is given in the interactive segmentation step.

All the above experiments exploit the Euclidean path length-based energy to compute the weight for the edge. This simple and intuitive way only considers the Euclidean length of the geodesic path in the graph construction. In this part, we utilize the geodesic distance map derived from the curvature-penalized metric~\cite{chen2017global} to estimate the cost weight, such that the image data and curvature information are introduced into the weight estimation. The FSRMetric model considering the curvature information combined with the adaptive contour scheme (FSRMetric-adpt) is utilized in the comparative experiment.
The qualitatively comparing results from the considered models on synthetic image, CT image and nature image are shown in Fig.~\ref{fig:CurvatureEnergy}. It can be observed that the proposed model can detect the target boundary correctly by utilizing the image data-based geodesic distance based energy.

Finally, we present examples for multiple object detection by the GroupPrps model in Fig.~\ref{fig:multiObjects}. The proposed method is a novel minimal path-based algorithm which works on simple closed contour with minimal interaction. The required internal point provides the priori knowledge about which of the object is to be detected. For multiple object detection, each of the target objects need a single user-provided point inside the target region. As shown in Fig.~\ref{fig:multiObjects}, the yellow dots represent the landmark points, to which the adaptive cuts associated are denoted as blue dash lines. The obtained closed contours for different objects are demonstrated with different color lines, illustrated in column $2$.

\section{Conclusions}
\label{sec:conclusion}
In this paper, we propose a new interactive image segmentation model partially  relying on a set of precomputed boundary proposals. A crucial point for the proposed model concentrates on computing the adaptive cut from the landmark point, yielding a disconnection constraint on the image domain. The final segmentation contour is regarded as an optimal circular path chosen from an admissible set, where each circular path is the concatenation of truncated boundary proposals and the corresponding connection paths. Accordingly, the proposed interactive segmentation model enables to blend the benefits from the prior information of the boundary proposals and the efficiency of the graph-based optimization scheme. The experimental results prove that the proposed models indeed outperform state-of-the-art minimal path-based interactive segmentation models.

Future work will be devoted to developing algorithms for automatic extraction of closed target boundaries based on the proposed model and the deep learning techniques. For instance, the proposed model requires a landmark point inside the target region to perform the model initialization. This landmark point can be either given by user, thus leading to an interactive segmentation method, or predicted  by a deep learning-based model, for instance the PolarMask model~\cite{xie2021polarmask++}.

\section*{Acknowledgment}
The authors would like to acknowledge the editors and the reviewers for their valuable time to improve the quality of this manuscript. This work is in part supported by the National Natural Science Foundation of China (NOs. 62102210, 62172243), the Shandong Provincial Natural Science Foundation (NOs.~ZR2021QF029, ZR2022YQ64), a project in QLUT (NO. 2022PY11), the Shanghai Sailing Program (NO.20YF1401500), the French government under management of Agence Nationale de la Recherche as part of the ``Investissements d'avenir'' program, reference ANR-19-P3IA-0001 (PRAIRIE 3IA Institute), and the Young Taishan Scholars under Grant (NO. tsqn201909137).

\appendix
\label{sec:appendix}
\subsection{Instances of Geodesic Metrics for Connection Paths and Edge Weights}
\label{subsec_Convexity}
In this appendix, we present two instances of the geodesic metrics for the computation of the connection paths between two adjacent boundary proposals as well as their connection cost (i.e. edge weights).  

\subsubsection{Isotropic Riemannian Metric}
The classical isotropic Riemannian metric has a simple form, which can be constructed using the potential $\psi_{\rm Seg}$ (see Eq.~\eqref{eq_SegPotential}). In this case, the isotropic metric $\cF:=\cF_{\rm IR}$ is defined over the space $\Omega\times\bR^2$ by setting $\bM=\Omega$ such that
\begin{equation}
\cF^{\rm IR}(\fx,\fu)=\psi_{\rm Seg}(\fx)\|\fu\|.
\end{equation}

\subsubsection{Curvature-penalized Metric}
The curvature regularization is able to impose rigid property to the segmentation contours, thus alleviating the unexpected influence from the image noise and spurious edges. For this purpose, we consider two curvature-penalized geodesic metrics, the Euler-Mumford elastica metric~\cite{chen2017global} or the Reeds-Shepp forward metric~\cite{duits2018optimal}. 

Let $\bS^1=[0,2\pi)$ be an interval with periodic boundary condition and let $\bM:=\Omega\times \bS^1\subset\bR^3$ be an orientation-lifted space. Both Euler-Mumford elastica and  Reeds-Shepp forward metrics are established over the space $\bM\times\bR^3\to[0,\infty]$. We respectively denote by $\cF^{\rm EM}$ and $\cF^{\rm RS}$ the Euler-Mumford elastica and  Reeds-Shepp forward metrics, which can be defined for any point $\tilde\fx=(\fx,\theta)\in\bM$ and any vector $\tilde\fu=(\fu,\nu)\in\bR^3$ such that
\begin{equation}
\cF^{\rm EM}(\tilde\fx,\tilde\fu)=	
\begin{cases}
\|\fu\|+\frac{\beta^2\nu^2}{\|\fu\|}, &\text{if~}\fu=\|\fu\|\fn_\theta\\
\infty, &\text{otherwise}
\end{cases}
\end{equation}
and
\begin{equation}
\cF^{\rm RS}(\tilde\fx,\tilde\fu)=	
\begin{cases}
\sqrt{\|\fu\|^2+\beta^2\nu^2}, &\text{if~}\fu=\|\fu\|\fn_\theta\\
\infty, &\text{otherwise}
\end{cases}
\end{equation}
where $\beta>0$ is weighting parameter and where $\fn_\theta=(\cos\theta,\sin\theta)$ is a uint vector associated to the ankle $\theta\in\bS^1$.

In the framework of the curvature-penalized models~\cite{chen2017global,duits2018optimal}, a  smooth curve $\gamma:[0,1]\to\Omega$ is lifted to an orientation-lifted curve $\tilde\gamma:=(\gamma,\eta):[0,1]\to\Omega\times\bS^1$ s.t. 
\begin{equation}
\label{eq_TurningAnkles}
\gamma^\prime(u)=\|\gamma^\prime(u)\|\fn_{\eta(u)},\quad \forall u\in[0,1].
\end{equation}
Note that an orientation-lifted curve $\tilde\gamma=(\gamma,\eta)$ satisfying Eq.~\eqref{eq_TurningAnkles} is comprised of two components, where $\gamma$ is its physical projection and $\eta:[0,1]\to\bS^1$ characterizes the ankles associated to the tangents $\gamma^\prime$.

Equation~\eqref{eq_TurningAnkles} also yields a new representation for the curvature $\kappa:[0,1]\to\bR$ of the smooth curve $\gamma$
\begin{equation}
\label{eq_CurvatureRep}
\kappa(u)=\frac{\eta^\prime(u)}{\|\gamma^\prime(u)\|}.	
\end{equation}
Thus the minimization of $\cL^{\rm EM}(\gamma)=\int_0^1(1+\beta^2\kappa^2)\|\gamma^\prime\|du$  and $\cL^{\rm RS}(\gamma)=\int_0^1\sqrt{1+\beta^2\kappa^2}\|\gamma^\prime\|du$ can be implemented by respectively minimizing the following energy functionals 
\begin{align}
&\cL_{\cF^{\rm EM}}(\tilde\gamma)=\int_0^1\cF^{\rm EM}(\tilde\gamma,\tilde\gamma^\prime)du\\
&\cL_{\cF^{\rm RS}}(\tilde\gamma)=\int_0^1\cF^{\rm RS}(\tilde\gamma,\tilde\gamma^\prime)du.
\end{align}
As introduced in~\cite{chen2017global,duits2018optimal,mirebeau2018fast} and analogous to the classical Cohen-Kimmel model, the minimization of the energy functionals $\cL_{\cF^{\rm EM}}$ and $\cL_{\cF^{\rm RS}}$ can be achieved by computing a geodesic distance map $\tilde\cD$, as the solution to the following PDE with $\cF:=\cF^{\rm EM}$ or $\cF:=\cF^{\rm RS}$, reading
\begin{equation}
\label{eq_OLPDE}
\max_{\tilde\fv\neq \mathbf{0}}\frac{\langle\nabla\tilde\cD(\tilde\fx),\tilde\fv \rangle}{\cF(\tilde\fx,\tilde\fv)}=1	
\end{equation}
with  boundary condition $\tilde\cD(\tilde\fx)=0,~\forall\tilde\fx\in\tilde\cS$, where $\tilde\cS\subset\bM$ is a connected subset. 

\noindent\emph{Application to Our Problem}. In Section~\ref{subsec_EdgeWeight}, a boundary proposal $\cS_i$ is a subset of the physical space $\Omega$. In order to apply the curvature-penalized metrics for estimating the connection paths and the edge weights, we should map each boundary $\cS_i$ to the orientation-lifted space $\bM$. For this purpose, let $\Gamma_i:[0,1]\to\cS_i$ be the parameterization of of $\cS_i$  and let $\eta_i:[0,1]\to\bS^1$ be a parametric function such that $\Gamma_i^\prime(u)=\|\Gamma_i^\prime(u)\|\fn_{\eta_i(u)}$. In this case, we obtain a series of orientation-lifted boundary proposals $\tilde\Gamma_i=(\Gamma_i,\eta_i)$. As a result, the geodesic distance map $\tilde\cD_{\tilde\Gamma_i}$ associated with $\tilde\Gamma_i$ can be obtained by solving the PDE~\eqref{eq_OLPDE} with boundary condition $\tilde\cD_{\tilde\Gamma_i}(\tilde\fx)=0,~\text{~if~}\exists u\in[0,1]\text{~such that~} \tilde\fx=\tilde\Gamma_i(u)$. The connection path $\tilde\cG_{i,j}:[0,1]\to\bM$ that links $\tilde\Gamma_i$ to $\tilde\Gamma_j$ can be obtained using the gradient descent ODE~\eqref{eq:GeoPathODE} and the distance map $\tilde\cD_{\tilde\Gamma_i}$. By the definition of the orientation lifting, $\tilde\cG_{i,j}$ can be rewritten as $\tilde\cG_{i,j}=(\cG_{i,j}, \eta_{i,j})$, where $\cG_{i,j}$ is the physical projection used for building the final segmentation contours. 

In addition, the cost $\kC_2$ in Eq.~\eqref{eq_CurvaEnergy} for edge weights can be reformulated as
\begin{align}
\kC_2(\cG_{i,j})&=\int_0^1\sqrt{1+\beta^2\kappa_{i,j}(u)^2}\|\cG_{i,j}^\prime(u)\|du\nonumber\\
	&=\int_0^1\sqrt{\|\cG_{i,j}^\prime(u)\|^2+\eta_{i,j}^\prime(u)^2}du,
\end{align}
which can be simultaneously computed during the estimation of the geodesic distances using the HFM, as  adopted in~\cite{chen2023Convexity}.

\bibliographystyle{IEEEbib}
\bibliography{pGTubular}

\begin{thebibliography}{10}

\bibitem{boykov2006graph}
Y.~Boykov and G.~Funka-Lea,
\newblock ``{Graph cuts and efficient ND image segmentation},''
\newblock {\em Int. J. Comput. Vis.}, vol. 70, no. 2, pp. 109--131, 2006.

\bibitem{vicente2008graph}
S.~Vicente, V.~Kolmogorov, and C.~Rother,
\newblock ``Graph cut based image segmentation with connectivity priors,''
\newblock in {\em Proc. CVPR}. IEEE, 2008, pp. 1--8.

\bibitem{couprie2010power}
C.~Couprie, L.~Grady, L.and~Najman, and H.~Talbot,
\newblock ``{Power watershed: A unifying graph-based optimization framework},''
\newblock {\em IEEE Trans. Pattern Anal. Mach. Intell.}, vol. 33, no. 7, pp.
  1384--1399, 2010.

\bibitem{grady2006random}
L.~Grady,
\newblock ``Random walks for image segmentation,''
\newblock {\em IEEE Trans. Pattern Anal. Mach. Intell.}, vol. 28, no. 11, pp.
  1768--1783, 2006.

\bibitem{yang2010user}
W.~Yang, J.~Cai, J.~Zheng, and J.~Luo,
\newblock ``User-friendly interactive image segmentation through unified
  combinatorial user inputs,''
\newblock {\em IEEE Trans. Image Process.}, vol. 19, no. 9, pp. 2470--2479,
  2010.

\bibitem{zhang2010diffusion}
J.~Zhang, J.~Zheng, and J.~Cai,
\newblock ``A diffusion approach to seeded image segmentation,''
\newblock in {\em Proc. CVPR}. IEEE, 2010, pp. 2125--2132.

\bibitem{li2004lazy}
Y.~Li, J.~Sun, C.-K. Tang, and H.-Y. Shum,
\newblock ``Lazy snapping,''
\newblock {\em ACM Trans. Graphics}, vol. 23, no. 3, pp. 303--308, 2004.

\bibitem{arbelaez2004energy}
P.~A. Arbel{\'a}ez and L.~D Cohen,
\newblock ``Energy partitions and image segmentation,''
\newblock {\em J. Math. Imag. Vis.}, vol. 20, no. 1, pp. 43--57, 2004.

\bibitem{bai2009geodesic}
X.~Bai and G.~Sapiro,
\newblock ``Geodesicmatting: A framework for fast interactive image and video
  segmentation and matting,''
\newblock {\em Int. J. Comput. Vis.}, vol. 82, no. 2, pp. 113--132, 2009.

\bibitem{price2010geodesic}
B.~L Price, B.~Morse, and S.~Cohen,
\newblock ``Geodesic graph cut for interactive image segmentation,''
\newblock in {\em Proc. CVPR}. IEEE, 2010, pp. 3161--3168.

\bibitem{chen2018fast}
D.~Chen and L.~D. Cohen,
\newblock ``Fast asymmetric fronts propagation for image segmentation,''
\newblock {\em J. Math. Imag. Vis.}, vol. 60, no. 6, pp. 766--783, 2018.

\bibitem{nguyen2012robust}
T.~N.~A. Nguyen, J.~Cai, J.~Zhang, and J.~Zheng,
\newblock ``Robust interactive image segmentation using convex active
  contours,''
\newblock {\em IEEE Trans. Image Process.}, vol. 21, no. 8, pp. 3734--3743,
  2012.

\bibitem{spencer2018parameter}
J.~Spencer, K.~Chen, and J.~Duan,
\newblock ``Parameter-free selective segmentation with convex variational
  methods,''
\newblock {\em IEEE Trans. Image Process.}, vol. 28, no. 5, pp. 2163--2172,
  2018.

\bibitem{gao20123d}
Y.~Gao, R.~Kikinis, S.~Bouix, M.~Shenton, and A.~Tannenbaum,
\newblock ``{A 3D interactive multi-object segmentation tool using local robust
  statistics driven active contours},''
\newblock {\em Med. Image Anal.}, vol. 16, no. 6, pp. 1216--1227, 2012.

\bibitem{wang2019deep}
G.~Wang, M.~A. Zuluaga, W.~Li, P.~Rosalind, P.~A. Patel, A.~Michael, A.~L.
  Divid, and O.~Sebastien,
\newblock ``{DeepIGeoS: A deep interactive geodesic framework for medical image
  segmentation},''
\newblock {\em IEEE Trans. Pattern Anal. Mach. Intell.}, vol. 41, no. 7, pp.
  1559--1572, 2019.

\bibitem{rother2004grabcut}
C.~Rother, V.~Kolmogorov, and A.~Blake,
\newblock ``{"GrabCut" interactive foreground extraction using iterated graph
  cuts},''
\newblock {\em ACM Trans. Graphics}, vol. 23, no. 3, pp. 309--314, 2004.

\bibitem{lempitsky2010image}
V.~Lempitsky, P.~Kohli, C.~Rother, and T.~Sharp,
\newblock ``Image segmentation with a bounding box prior,''
\newblock in {\em Proc. ICCV}, 2010.

\bibitem{kass1988snakes}
M.~Kass, A.~Witkin, and D.~Terzopoulos,
\newblock ``{Snakes: Active contour models},''
\newblock {\em Int. J. Comput. Vis.}, vol. 1, no. 4, pp. 321--331, 1988.

\bibitem{xu1998snakes}
C.~Xu and J.~L. Prince,
\newblock ``Snakes, shapes, and gradient vector flow,''
\newblock {\em IEEE Trans. Image Process.}, vol. 7, no. 3, pp. 359--369, 1998.

\bibitem{li2008minimization}
C.~Li, C.-Y. Kao, J.~C Gore, and Z.~Ding,
\newblock ``Minimization of region-scalable fitting energy for image
  segmentation,''
\newblock {\em IEEE Trans. Image Process.}, vol. 17, no. 10, pp. 1940--1949,
  2008.

\bibitem{brox2009local}
T.~Brox and D.~Cremers,
\newblock ``{On local region models and a statistical interpretation of the
  piecewise smooth Mumford-Shah functional},''
\newblock {\em Int. J. Comput. Vis.}, vol. 84, no. 2, pp. 184--193, 2009.

\bibitem{cai2021avlsm}
Q.~Cai, Y.~Qian, S.~Zhou, J.~Li, Y.-H. Yang, F.~Wu, and D.~Zhang,
\newblock ``{AVLSM: Adaptive variational level set model for image segmentation
  in the presence of severe intensity inhomogeneity and high noise},''
\newblock {\em IEEE Trans. Image Process.}, vol. 31, pp. 43--57, 2022.

\bibitem{badshah2010image}
N.~Badshah and K.~Chen,
\newblock ``Image selective segmentation under geometrical constraints using an
  active contour approach,''
\newblock {\em Commun. Comput. Phys.}, vol. 7, no. 4, pp. 759, 2010.

\bibitem{mabood2016active}
L.~Mabood, H.~Ali, N.~Badshah, K.~Chen, and G.~A. Khan,
\newblock ``Active contours textural and inhomogeneous object extraction,''
\newblock {\em Pattern Recognit.}, vol. 55, pp. 87--99, 2016.

\bibitem{windheuser2009beyond}
T.~Windheuser, T.and~Schoenemann and D.~Cremers,
\newblock ``{Beyond connecting the dots: A polynomial-time algorithm for
  segmentation and boundary estimation with imprecise user input},''
\newblock in {\em Proc. ICCV}. IEEE, 2009, pp. 717--722.

\bibitem{miranda2012riverbed}
P.~AV Miranda, A.~X. Falcao, and T.~V. Spina,
\newblock ``{Riverbed: A novel user-steered image segmentation method based on
  optimum boundary tracking},''
\newblock {\em IEEE Trans. Image Process.}, vol. 21, no. 6, pp. 3042--3052,
  2012.

\bibitem{mortensen1998interactive}
E.~Mortensen and W.~Barrett,
\newblock ``Interactive segmentation with intelligent scissors,''
\newblock {\em Graph. Models and Image Process.}, vol. 60, no. 5, pp.
  349–384, 1998.

\bibitem{dijkstra1959note}
E.~W. Dijkstra,
\newblock ``A note on two problems in connexion with graphs,''
\newblock {\em Numer. Math.}, vol. 1, no. 1, pp. 269--271, 1959.

\bibitem{cohen1997global}
L.~D. Cohen and R.~Kimmel,
\newblock ``{Global minimum for active contour models:~A minimal path
  approach},''
\newblock {\em Int. J. Comput. Vis.}, vol. 24, no. 1, pp. 57--78, 1997.

\bibitem{peyre2010geodesic}
G.~Peyr{\'e}, M.~P{\'e}chaud, R.~Keriven, and L.~D Cohen,
\newblock ``Geodesic methods in computer vision and graphics,''
\newblock {\em Foundations and Trends{\textregistered} in Computer Graphics and
  Vision}, vol. 5, no. 3--4, pp. 197--397, 2010.

\bibitem{benmansour2009fast}
F.~Benmansour and L.~D. Cohen,
\newblock ``{Fast object segmentation by growing minimal paths from a single
  point on 2D or 3D images},''
\newblock {\em J. Math. Imag. Vis.}, vol. 33, no. 2, pp. 209--221, 2009.

\bibitem{mille2015combination}
J.~Mille, S.~Bougleux, and L.~D. Cohen,
\newblock ``Combination of piecewise-geodesic curves for interactive image
  segmentation,''
\newblock 2015, vol. 112, p. 1–22.

\bibitem{chen2017global}
D.~Chen, J.-M. Mirebeau, and L.~D. Cohen,
\newblock ``{Global minimum for a Finsler elastica minimal path approach},''
\newblock {\em Int. J. Comput. Vis.}, vol. 122, no. 3, pp. 458--483, 2017.

\bibitem{cohen2001multiple}
L.~D. Cohen,
\newblock ``Multiple contour finding and perceptual grouping using minimal
  paths,''
\newblock {\em J. Math. Imaging Vis.}, vol. 14, no. 3, pp. 225--236, 2001.

\bibitem{chen2019region}
D.~Chen, J.-M. Mirebeau, H.~Shu, and L.~D. Cohen,
\newblock ``A region-based randers geodesic approach for image segmentation,''
\newblock {\em arXiv preprint arXiv:1912.10122}, 2019.

\bibitem{appleton2005globally}
B.~Appleton and H.~Talbot,
\newblock ``Globally optimal geodesic active contours,''
\newblock {\em J. Math. Imag. Vis.}, vol. 23, no. 1, pp. 67--86, 2005.

\bibitem{chen2021geodesic}
D.~Chen, J.~Zhu, X.~Zhang, M.~Shu, and L.~D. Cohen,
\newblock ``Geodesic paths for image segmentation with implicit region-based
  homogeneity enhancement,''
\newblock {\em IEEE Trans. Image Process.}, vol. 30, pp. 5138--5153, 2021.

\bibitem{wang2013interactive}
L.~Wang, V.~Kallem, M.~Bansal, J.~Eledath, H.~Sawhney, K.~Karp, D.~J Pearson,
  M.~D Mills, G.~E Quinn, and R.~A Stone,
\newblock ``Interactive retinal vessel extraction by integrating vessel tracing
  and graph search,''
\newblock in {\em Proc. MICCAI}, 2013, pp. 567--574.

\bibitem{liu2021new}
L.~Liu, D.~Chen, M.~Shu, H.~Shu, and L.~D. Cohen,
\newblock ``A new tubular structure tracking algorithm based on
  curvature-penalized perceptual grouping,''
\newblock in {\em Proc. ICASSP}, 2021, pp. 2195--2199.

\bibitem{liuTractory2022}
L.~Liu, D.~Chen, M.~Shu, B.~Li, H.~Shu, M.~Paques, and L.~D. Cohen,
\newblock ``Trajectory grouping with curvature regularization for tubular
  structure tracking,''
\newblock {\em IEEE Trans. Image Process.}, vol. 31, pp. 405--418, 2022.

\bibitem{sethian1999fast}
J.~A. Sethian,
\newblock ``Fast marching methods,''
\newblock {\em SIAM Review}, vol. 41, no. 2, pp. 199--235, 1999.

\bibitem{sochen1998general}
N.~Sochen, R.~Kimmel, and R.~Malladi,
\newblock ``A general framework for low level vision,''
\newblock {\em IEEE Trans. Image Process.}, vol. 7, no. 3, pp. 310–318, 1998.

\bibitem{canny1986computational}
J.~Canny,
\newblock ``A computational approach to edge detection,''
\newblock {\em IEEE Trans. Pattern Anal. Mach. Intell.}, , no. 6, pp. 679--698,
  1986.

\bibitem{arbelaezContour2011}
P.~Arbel\'aez, M.~Maire, C.~Fowlkes, and J.~Malik,
\newblock ``Contour detection and hierarchical image segmentation,''
\newblock {\em IEEE Trans. Pattern Anal. Mach. Intell.}, vol. 33, no. 5, pp.
  898--916, 2011.

\bibitem{jacob2004design}
M.~Jacob and M.~Unser,
\newblock ``Design of steerable filters for feature detection using canny-like
  criteria,''
\newblock {\em IEEE Trans. Pattern Anal. Mach. Intell.}, vol. 26, no. 8, pp.
  1007--1019, 2004.

\bibitem{duits2018optimal}
R.~Duits, S.~PL Meesters, J.-M. Mirebeau, and J.~M Portegies,
\newblock ``{Optimal paths for variants of the 2D and 3D Reeds--Shepp car with
  applications in image analysis},''
\newblock {\em J. Math. Imag. Vis.}, vol. 60, no. 6, pp. 816--848, 2018.

\bibitem{mirebeau2018fast}
J.-M. Mirebeau,
\newblock ``Fast-marching methods for curvature penalized shortest paths,''
\newblock {\em J. Math. Imag. Vis.}, vol. 60, no. 6, pp. 784--815, 2018.

\bibitem{mirebeau2019riemannian}
J.-M. Mirebeau,
\newblock ``{Riemannian fast-marching on Cartesian grids, using Voronoi's first
  reduction of quadraticforms},''
\newblock {\em SIAM J. Numer. Anal.}, vol. 57, no. 6, pp. 2608--2655, 2019.

\bibitem{mirebeau2021massively}
J.-M. Mirebeau, L.~Gayraud, R.~Barr{\`e}re, D.~Chen, and F.~Desquilbet,
\newblock ``Massively parallel computation of globally optimal shortest paths
  with curvature penalization,''
\newblock {\em Concurrency Computat., Pract. Exper.}, p. e7472, 2021.

\bibitem{xie2021polarmask++}
E.~Xie, W.~Wang, M.~Ding, R.~Zhang, and P.~Luo,
\newblock ``{Polarmask++: Enhanced polar representation for single-shot
  instance segmentation and beyond},''
\newblock {\em IEEE Trans. Pattern Anal. Mach. Intell.}, vol. 44, no. 9, pp.
  5385--5400, 2021.

\bibitem{chen2023Convexity}
D.~Chen, J.-M. Mirebeau, M.~Shu, X.~Tai, and L.~D. Cohen,
\newblock ``Geodesic models with convexity shape prior,''
\newblock {\em IEEE Trans. Pattern Anal. Mach. Intell.}, vol. 45, no. 7, pp.
  8433--8452, 2023.

\end{thebibliography}

\ifCLASSOPTIONcaptionsoff
  \newpage
\fi

\end{document}